\documentclass[sigconf]{acmart}

\usepackage[ruled]{algorithm2e}
\usepackage{multirow}
\usepackage{subfigure}
\usepackage{enumitem}
\usepackage{sistyle}
\SIthousandsep{,}
\usepackage{makecell}
\usepackage{multirow}
\usepackage{enumitem}
\usepackage{filecontents}
\usepackage{bbm}
\usepackage{pgfplots}
\usetikzlibrary{patterns}

\AtBeginDocument{%
  \providecommand\BibTeX{{%
    \normalfont B\kern-0.5em{\scshape i\kern-0.25em b}\kern-0.8em\TeX}}}

\copyrightyear{2022} 
\acmYear{2022} 
\setcopyright{acmcopyright}
\acmConference[CIKM '22]{Proceedings of the 31st ACM International Conference on Information and Knowledge Management}{October 17--21, 2022}{Atlanta, GA, USA}
\acmBooktitle{Proceedings of the 31st ACM International Conference on Information and Knowledge Management (CIKM '22), October 17--21, 2022, Atlanta, GA, USA}
\acmPrice{15.00}
\acmDOI{10.1145/3511808.3557271}
\acmISBN{978-1-4503-9236-5/22/10}

\begin{document}

\title{CorpusBrain: Pre-train a Generative Retrieval Model for Knowledge-Intensive Language Tasks}

\author{Jiangui Chen}
\affiliation{
	\institution{CAS Key Lab of Network Data Science and Technology, ICT, CAS}
	\institution{University of Chinese Academy of Sciences}
	\city{Beijing}
	\country{China}
}
\email{chenjiangui18z@ict.ac.cn}
 
\author{Ruqing Zhang}
\affiliation{
	\institution{CAS Key Lab of Network Data Science and Technology, ICT, CAS}
	\institution{University of Chinese Academy of Sciences}
	\city{Beijing}
	\country{China}
}
\email{zhangruqing@ict.ac.cn}
 
\author{Jiafeng Guo}
\authornote{Jiafeng Guo is the corresponding author.}
\affiliation{
	\institution{CAS Key Lab of Network Data Science and Technology, ICT, CAS}
	\institution{University of Chinese Academy of Sciences}
	\city{Beijing}
	\country{China}
}
\email{guojiafeng@ict.ac.cn}

\author{Yiqun Liu}
\affiliation{
	\institution{BNRist, DCST, Tsinghua University}
	\city{Beijing}
	\country{China}
}
\email{yiqunliu@tsinghua.edu.cn}

\author{Yixing Fan}
\affiliation{
	\institution{CAS Key Lab of Network Data Science and Technology, ICT, CAS}
	\institution{University of Chinese Academy of Sciences}
	\city{Beijing}
	\country{China}
}
\email{fanyixing@ict.ac.cn}
 
\author{Xueqi Cheng}
\affiliation{
	\institution{CAS Key Lab of Network Data Science and Technology, ICT, CAS}
	\institution{University of Chinese Academy of Sciences}
	\city{Beijing}
	\country{China}
}
\email{cxq@ict.ac.cn}

\renewcommand{\shortauthors}{}


\begin{abstract}

Knowledge-intensive language tasks (KILT) usually require a large body of information to provide correct answers. 
A popular paradigm to solve this problem is to combine a search system with a machine reader, where the former retrieves supporting evidences and the latter examines them to produce answers. 
Recently, the reader component has witnessed significant advances with the help of large-scale pre-trained generative models. Meanwhile most existing solutions in the search component rely on the traditional ``index-retrieve-then-rank'' pipeline, which suffers from large memory footprint and difficulty in end-to-end optimization. 
Inspired by recent efforts in constructing model-based IR models, we propose to replace the traditional multi-step search pipeline with a novel single-step generative model, which can dramatically simplify the search process and be optimized in an end-to-end manner. 
We show that a strong generative retrieval model can be learned with a set of adequately designed pre-training tasks, and be adopted to improve a variety of downstream KILT tasks with further fine-tuning. 
We name the pre-trained generative retrieval model as \textit{CorpusBrain} as all information about the corpus is encoded in its parameters without the need of constructing  additional index. 
Empirical results show that CorpusBrain can significantly outperform strong baselines for the retrieval task on the KILT benchmark and establish new state-of-the-art downstream performances. We also show that CorpusBrain works well under zero- and low-resource settings.

\end{abstract}

\ccsdesc[500]{Information systems~Retrieval models and ranking}

\keywords{Model-based IR; Pre-training; Generative Retrieval}

\maketitle

\section{Introduction}

Knowledge-intensive language tasks (KILT) such as fact checking~\cite{fever} and open-domain question answering~\cite{hotpotqa}, have received much attention in recent years. 
Compared with traditional information processing tasks, they are usually more complex and require to surface knowledge from a large body of information. 
A popular paradigm to approach these tasks is to combine a search system with a machine reader \cite{kilt}. 
The former retrieves a limited subset of supporting evidences from large corpora, and the latter then examines the retrieved information to produce final answers. 
Recently, the reader component has witnessed significant advances with the help of the large-scale pre-trained generative models (e.g., BART~\cite{bart} and T5~\cite{bart}), and such models have become the de-facto implementation of the reader.  
However, the search component has yet to benefit from these tremendous advances in generative models.

\begin{figure*}[t]
 \centering
 \includegraphics[scale=0.73]{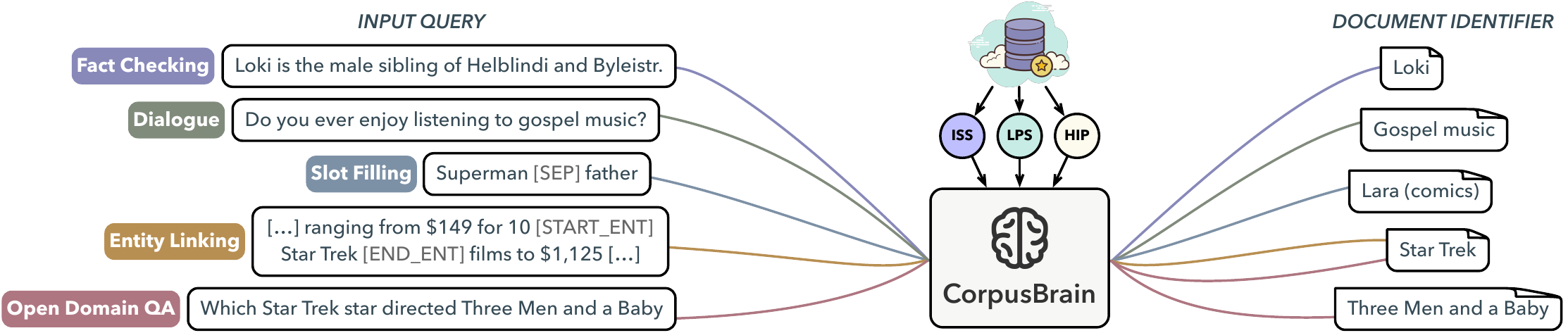}
 \caption{Overview of how the CorpusBrain can be adopted to solve a variety of downstream KILT tasks. At the pre-training stage, we design three pre-training tasks (i.e., ISS, LPS and HIP) to encode the knowledge about the corpus, as shown in Figure \ref{fig:tasks}. If downstream supervised data is available, CorpusBrain can be further fine-tuned to improve the retrieval performance. Given a query, the retrieval task is cast as a Seq2Seq learning problem, to generate the identifiers of relevant documents.}
 \label{fig:intro}
\end{figure*}

Most existing solutions in the search component follow the paradigm of  ``index-retrieve-then-rank'' \cite{deepct, zhan2021optimizing, dpr, zheng2015learning, frej2020learning} which includes  three sequential steps: (1) building an index for each document in the corpus; (2) retrieving an initial set of candidate documents for a query; and (3) determining the relevance degree of each candidate. 
Despite its wide usage, this paradigm has clear limitations. 
At the training stage, heterogeneous ranking components are usually difficult to be optimized in an end-to-end way towards the global objective. 
At the inference stage, a large document index is needed to search over the corpus, leading to significant memory consumption and computational overhead. Besides, errors would accumulate and propagate among the sequential components.

Recently,  \citet{metzler2021rethinking} envisioned a fundamentally different paradigm called model-based IR, to replace the long-standing  ``index-retrieve-then-rank'' paradigm. Specifically, with model-based IR, the indexing, retrieval, and ranking components of traditional IR systems are collapsed into a single consolidated model.   
This enables us to mitigate the aforementioned technical issues. Firstly, the knowledge of all documents in the corpus is encoded into the model parameters, which can be optimized directly in an end-to-end manner.  
Secondly, the memory and computational cost is greatly reduced because the document index is eliminated. Documents are returned using model parameters only, which can dramatically simplify the heavy search  process. 
Inspired by this blueprint, researchers  \cite{genre,gere,tay2022transformer} have proposed to generate identifier strings for documents via generative language models. 
A typical way is to directly fine-tune the off-the-shelf pre-trained generative models (e.g., BART~\cite{bart}) for specific KILT task, e.g., entity linking \cite{genre} and fact checking \cite{gere}. 
Unfortunately, these methods depend on large-scale application-specific supervision, which are often not available for many real-world applications. 

Therefore, in this paper, we propose to pre-train a novel single-step  generative model, called \textit{CorpusBrain}, to encode all information of the corpus within its parameters in a general way. 
In this work, we assume that the pre-training data is defined as positive pairs of query and document identifier.   
To resemble the relevance relationship between query and document in downstream KILT tasks, the pre-training task should be designed to meet the following requirements: 
(R1) It should capture different granularities of semantics between the query and document. For example, the input queries of different KILT tasks range from sentence-level \cite{fever,nq, hotpotqa, triviaqa, eli5} to paragraph-level \cite{aida, wned,wow}.   
(R2) It should be flexible to predict dynamic numbers of relevant documents for different queries. For example, the question answering usually requires multiple documents to support the answer, while the slot filling task \cite{zsre, trex} usually needs one. 
(R3) It should capture inter-document semantic relation since multiple documents may share similar characteristics with respect to a query.  

In light of the above requirements, we carefully devise three pre-training tasks to capture the query-document relevance in different views, namely Inner Sentence Selection (ISS), Lead Paragraph Selection (LPS), and Hyperlink Identifier Prediction (HIP). 
The key idea of ISS and LPS is to sample sentences or paragraphs from documents and adopt them as pseudo queries, while that of HIP is to utilize hyperlinks and anchor texts to  approximate the relationship between two documents.  
Based on the three proposed tasks, we can build large-scale pseudo pairs of query and document identifiers without additional human supervision. 
Then, initialized with BART \cite{bart}, we pre-train our model via a standard seq2seq objective, i.e., maximizing the likelihood of the output sequence with teacher forcing \cite{sutskever2011generating}. 
As shown in Figure \ref{fig:intro}, once such a strong generative retrieval model is learned, we expect it can be seamlessly adapted to a wide variety of downstream KILT tasks without the need of additional index. 


We pre-train CorpusBrain on English Wikipedia, which contains tens of millions of well-formed Wiki articles. 
We fine-tune CorpusBrain on the comprehensive KILT benchmark \cite{kilt} which consists of eleven datasets spanning five distinct KILT tasks. 
Empirical experimental results demonstrated that CorpusBrain can achieve significant improvements over strong baseline solutions for the retrieval task and push forward the SOTA performance on  a number of downstream tasks. 
We simulate both zero- or low-resource settings and show that CorpusBrain works well even when they were fine-tuned with very little supervision.

\section{Related Work}
In this section, we briefly review three lines of the related works, including traditional pipeline IR framework, model-based IR approaches and knowledge-intensive language tasks.

\vspace*{-2mm}
\subsection{Traditional Pipeline IR Framework}
Most existing IR methods follow a common thee-step pipeline framework, i.e., ``index-retrieve-then-rank''. 
For the indexing and retrieval stage, existing approaches can be divided into two categories~\cite{guo2022semantic} from the view of representation type and index mode, including sparse retrieval and dense retrieval models. 
For sparse retrieval, they generally build the inverted index based on the corpus, which encode term-based features like  term frequencies and term position.
The typical methods in this category are TF-IDF and BM25~\cite{bm25}. 
Besides, several works~\cite{zheng2015learning, frej2020learning} employed word embedding to enhance the semantic matching.
With the development of pre-training techniques, researchers  explore to utilize pre-trained models to estimate term weights. For example, DeepCT~\cite{deepct} used BERT~\cite{bert} to obtain term weights for the inverted index. 
For dense retrieval, they usually project documents into dense representations to build index and turn to approximate nearest neighbor search algorithms for fast retrieval. 
\citet{dpr} demonstrated dense retrieval models could outperform BM25 by utilizing in-batch negatives.
Recently, various fine-tuning techniques~\cite{xiong2020approximate, zhan2021optimizing, khattab2020colbert, hofstatter2021efficiently,zhan2021optimizing,ma2022pre} are explored to enhance dense retrieval.

For the ranking stage, many different ranking models have been proposed, including vector space models~\cite{salton1975vector}, probabilistic models~\cite{robertson1976relevance}, learning to rank models~\cite{liu2009learning, li2014learning, burges2006learning} and neural ranking models~\cite{guo2016deep, dai2018convolutional}.
Recently, researcher have shown that developed pre-training objectives tailed for IR could further enhance the performance on downstream ranking tasks~\cite{prop, bprop, ma2021pre, chang2019pre, lee2019latent}.
For example, ~\citet{lee2019latent} introduced Inverse Cloze Task (ICT) for passage retrieval by randomly sampling a sentence from passage as pseudo query and taking the rest sentences as the document.
~\citet{prop} presented Representative Words Prediction (ROP) task for pre-training, which sampled representation words from the document according to a unigram document language model. 
~\citet{ma2021pre} proposed HARP by leveraging the large-scale hyperlinks and anchor texts to pre-train the language model for ad-hoc retrieval. 
Despite their success, however, such pipeline framework has has drawbacks on the end-to-end optimization and memory resources.

\subsection{Model-based IR Approaches}
To replace the long-lived ``index-retrieve-then-rank'' paradigm, the model-based IR paradigm is proposed to collapse the indexing, retrieval, and ranking components of traditional IR systems into a single consolidate model \cite{metzler2021rethinking}. 
There have been some preliminary explorations in model-based IR~\cite{genre, gere, tay2022transformer,zhou2022dynamicretriever, seal} over the past year. 
For example, ~\citet{genre} presented to generate entity names in an autoregressive fashion for entity linking task.
~\citet{gere} proposed GERE for fact checking task, which generates the document titles as well as evidence sentence identifiers. 
~\citet{tay2022transformer} explored different ways to obtain  document identifiers, including unstructured atomic identifiers and semantically structured identifiers. 
However, these approaches are generally designed for specific  task, resulting low flexibility for other tasks. In addition, a large number of annotated data is needed to learn a well behaved model. 
In this work, we propose to pre-train a general-purpose generative model which can serve a wide range of downstream KILT  tasks.

\subsection{Knowledge-Intensive Language Tasks}
Knowledge-intensive language tasks (KILT) require access to large and external knowledge sources. 
For example, fact checking requires to find trustworthy evidences to determine the veracity of a claim \cite{fever}. 
Open-domain question answering needs to reason over a knowledge source to produce the correct answer for a question \cite{nq, hotpotqa, triviaqa, eli5}. 
Practical solutions to these tasks usually apply a two-step pipeline framework~\cite{fever, drqa}. 
First, an efficient retrieval component is employed to retrieve relevant information from a large knowledge source.
Then, dedicated downstream models are adopted to produce the final results by capturing the relationship between the input and the context of the retrieved information.

Up to now, numerous datasets for KILT tasks~\cite{fever, nq, hotpotqa, wow} have been proposed to facilitate the research.
Generally, these datasets have different formats, and are processed or evaluated with different assumptions. 
Besides, their knowledge sources vary from different versions of Wikipedia to entirely different corpora. 
To facilitate the task-to-task comparisons, a benchmark named KILT~\cite{kilt} was introduced, which consists of eleven datasets spanning five distinct tasks (i.e., fact checking, open domain question answering, slot filling, entity linking and dialogue). 
Crucially, all tasks in KILT are formulated into a common interface and grounded in the same snapshot of Wikipedia.

\section{Our Approach}
In this section, we present the novel pre-trained generative retrieval model CorpusBrain in detail. 
We first introduce our motivation on the design of our method. 
We then describe the model architecture as well as the pre-training tasks.

\begin{figure*}[t]
 \centering
 \includegraphics[scale=0.38]{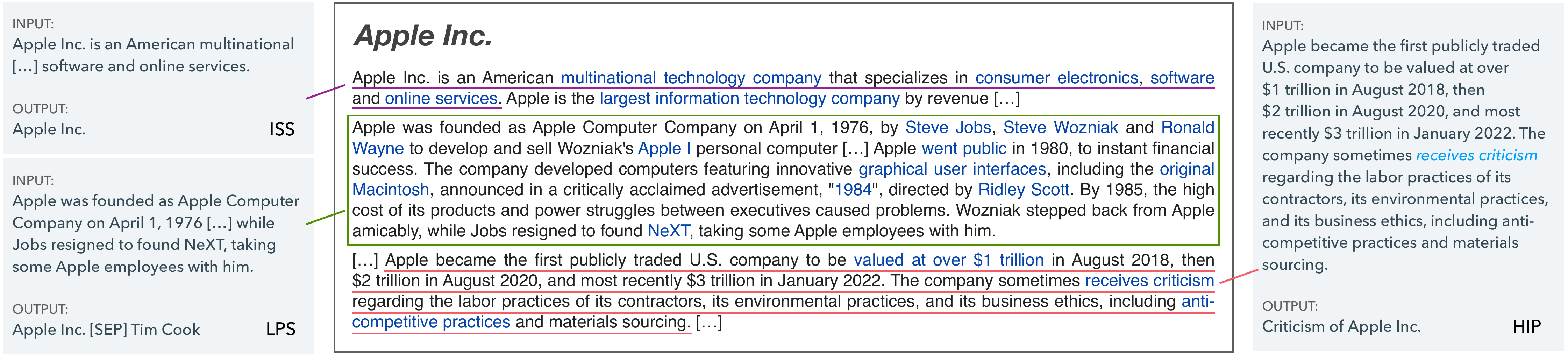}
 \caption{An illustrative example of the three pre-training tasks where each query is highlighted in different colors. Given a document $p_i$, the input of the ISS is a randomly sampled sentence from $p_i$, while the input of the LPS is a paragraph  from the first few paragraphs in $p_i$. The output of both the ISS and LPS is the document identifier of $p_i$ and destination pages linked by $o$ anchor texts sampled from $a_i$. For the HIP, the input is the contextual information of an anchor $a_i^k$ while the output is the document identifier of the destination page.}
 \label{fig:tasks}
\end{figure*}
\vspace*{-2mm}
\subsection{Motivation}

KILT tasks are usually approached by combining a search component with a reader component \cite{kilt}. 
Recently, thanks to the popularity of pre-trained models and their strong natural language understanding capabilities, the large-scale pre-trained generative models have been the de-facto implementation of the reader component. 
Unfortunately, these tremendous advances in pre-trained generative models has yet to bring similar transformational changes in how the search component is approached. 
Most existing solutions in the search component follow the traditional ``index-retrieve-then-rank'' paradigm.
To fully parameterize the traditional pipeline framework, ~\citet{metzler2021rethinking} outlined a high-level vision of the next generation of IR systems, called model-based IR, to  replace the long-lived pipeline framework with a single consolidated model. 
In this way, we can not only optimize the entire model directly in an end-to-end way, but also consume largely reduced memory resources and computation cost.

Motivated by this blueprint, there have been some preliminary works ~\cite{genre, gere, zhou2022dynamicretriever, tay2022transformer} dedicated to mapping a query to a document identifier for the search component.  
Typically, these solutions directly fine-tune the off-the-shelf pre-trained generative models (e.g., BART) on specific downstream  KILT tasks.
However, they are designed to suit the need of specific tasks. It may not be practical to deploy separate specialised models for each application due to considerable memory resources or computation overhead. 
Besides, a large amount of application-specific annotated data is required to learn the model, which is unsuitable for low data regimes, e.g., zero and few-shot settings. 
 
Therefore, in this work, we propose to pre-train a general-purpose generative retrieval model, named \textit{CorpusBrain}, by encoding all information about the corpus in the parameters. 
Namely, we target a strong generative retrieval model learned with several pre-training tasks, that can be used for a diverse range of KILT tasks without the need of additional index.  
Specifically, we carefully design three self-supervised pre-training objectives to satisfy the following three properties: (R1) The semantic granularities between the query and document vary greatly in different tasks; 
(R2) Various tasks usually require different numbers of retrieved supporting documents; 
(R3) Different documents could share similar characteristics with the queries. 
Then, we pre-train an encoder-decoder architecture to generate document identifiers via a standard seq2seq objective.

\vspace*{-2mm}
\subsection{Model Architecture}

To tackle the generative retrieval problem, we leverage a Transformer-based encoder-decoder architecture, which contains the following two dependent components: 
(1) Query Encoder, a bidirectional encoder to achieve the query representation;
(2) Identifier Decoder, a sequential generation process to yield document identifiers.

\subsubsection{Query Encoder}
The query encoder is to map the input query $q = \{w_1, w_2, \dots, w_{\vert q  \vert}\}$ into a compact vector that can capture its essential topics.
Specifically, the encoder represents the query $q$ as a series of hidden vectors, i.e., 
\begin{equation*}
    H_q = \text{Encoder}(w_1, w_2, \dots, w_{\vert q \vert}),
\end{equation*}
where $H_q$ denotes the query representation.  

\subsubsection{Identifier Decoder}
The decoder is to generate a sequence of document identifiers of the relevant documents for each query. 
Note the document identifier can be defined in different ways, such as the unique title and url of the document. 
Specifically, the probability of generating the $n$-th token $w_{m, n}$ in the $m$-th document identifier $t_m$ is defined as, 
\begin{equation*}
    p(w_{m,n}|w_{\leq m, <n}, q) = \text{Decoder}(w_{\leq m, <n}, H_{q}). 
\end{equation*}

At the inference time, we apply the constrained beam search strategy~\cite{genre} to limit each generated identifier to be in the valid pre-defined candidate set, i.e., the identifiers of all the documents in the given corpus. 
Concretely, we define our constrain in terms of a prefix tree where nodes are annotated with tokens from the predefined candidate set.
For each node in the tree, its children indicate all the valid continued token list from the prefix defined traversing the tree from the root to it.

\vspace*{-2mm}
\subsection{Pre-training Tasks}

Formally, suppose $\mathcal{C}=\{p_0, p_1, \dots\}$ denotes a large-scale corpus where $p_i$ denotes an individual document. 
Each document $p_i=\{t_i, s_i, a_i\}$ consists of a unique document identifier $t_i$, a sequence of sentences $s_i$, and an anchor set $a_i$. 
Specifically, $s_i=\{s_i^0, s_i^1, \dots, s_i^n\}$ contains $n$ sentences where $s_i^j$ denotes the $j$-th sentence in $p_i$. 
$a_i=\{a_i^0, a_i^1, \dots, a_i^m\}$ contains $m$ anchors where $a_i^k$ is the $k$-th anchor in $p_i$. 
$t_{a_i} = \{t_{a_i}^0, t_{a_i}^1, \dots, t_{a_i}^m\}$ are the  document identifiers of destination pages linked by each anchor text in $a_i$, where $t_{a_i}^l$ is linked by $a_i^l$.

What pre-training tasks are useful for improving Transformer-based encoder-decoder models is a crucial step in large-scale generative retrieval. 
It is generally hypothesized that using a pre-training task that more closely resembles the downstream task contributes to better fine-tuning performance \cite{zhang2020pegasus}. 
To satisfy three requirements discussed above, we first assume that the pre-training data is defined as positive pairs of query and document identifier.
Specifically, the document identifier can be defined in different ways, such as the unique title or url of the document. 
Then, as shown in Figure~\ref{fig:tasks}, we carefully design three pre-training tasks, i.e., Inner Sentence Selection (ISS), Lead Paragraph Selection (LPS), and Hyperlink Identifier Prediction (HIP). 
These three tasks target to learn the query-document relevance in different views, to generate pseudo pairs of query and document identifier to simulate the downstream retrieval task.  
The training data for these tasks can be freely obtained from the given corpus without an additional manual labeling process. 
In the following, we will present the proposed three pre-training tasks in detail.

\subsubsection{Inner Sentence Selection (ISS)}

Given a document $p_i$, to satisfy the requirement (R1) where the query for many KILT tasks (e.g., fact checking and question answering) is in the sentence level, the ISS treats the inner sentence $s_i^j$ randomly drawn from $p_i$ as the pseudo query $q$. 
This task contributes to capturing the semantic context of a sentence. 
To satisfy the requirement (R2), the ISS treats the $p_i$ and destination pages linked by $o (o < m)$ anchor texts $\{a_i^{s_1},a_i^{s_2},\dots,a_i^{s_o}\}$ randomly sampled from $a_i$ as the relevant documents.
ISS uses the concatenated document identifiers as the generation target [$t_i, t_{a_i}^{s_1}, t_{a_i}^{s_2},\dots,  t_{a_i}^{s_o}$]. 
In this way, it is feasible to achieve dynamic predictions of relevant documents for different downstream tasks. 

\subsubsection{Lead Paragraph Selection (LPS)}

Different from ISS, LPS samples a paragraph from the document $p_i$ and adopts it as the pseudo query $q$. 
With such pre-training task, we can fit the requirement (R1) where the query of entity linking, dialog and other KILT tasks is in the paragraph level.   
Generally, the top paragraphs can be a surrogate summary of better quality than the remains. 
Inspired by this, we regard the leading $l$ paragraphs as the pseudo queries. 
To satisfy the requirement (R2), the output [$t_i, t_{a_i}^{s_1}, t_{a_i}^{s_2},\dots,  t_{a_i}^{s_o}$] of the LPS is consistent with that of the ISS.

\subsubsection{Hyperlink Identifier Prediction (HIP)}

Based on the classical anchor intuition \cite{ma2021pre}, the hyperlink information could indicate the inter-document semantic relation to some extent. 
To satisfy the requirement (R3), we first view the anchor text  as a pseudo query. 
Unfortunately, the anchor texts are usually too short to carry enough semantics. 
Therefore, we resort to the anchor context, i.e., the surrounding contextual information in the anchor's corresponding sentence. 
Specifically, given a page $p_i$, we randomly select an anchor $a_i^k$ from the anchor set $a_i$ and locate the sentence $s_q$ containing $a_i^k$. 
Based on $s_q$, we get the anchor context by looking at its previous and successive sentence, i.e., $[s_{q-1}, s_q, s_{q+1}]$ as the pseudo query $q$. 
Then, the generation target is the document identifier $t_{a_i}^k$ of the destination page linked by the anchor $a_i^k$.

\begin{table}[t]
\small
    \centering
    \renewcommand{\arraystretch}{0.9}
    \setlength\tabcolsep{2.7pt}
    \caption{Retrieval datasets statistics of the KILT benchmark. \#P denotes the average number of relevant Wikipedia pages for each query in the train, dev and test set. `-' denotes that the task does not provide a ground-truth document in the training set.}
    \label{tab:datasets}
    \begin{tabular}{lccrrr}
        \toprule
        \textbf{Dataset}  & \textbf{Task} & \textbf{\#P} & \textbf{Train Size} & \textbf{Dev Size} & \textbf{Test Size} \\
        \midrule
        \textbf{FEV}~\cite{fever} & Fact Checking & 1.13 & 104,966 & 10,444 & 10,100 \\
        \textbf{AY2}~\cite{aida} & Entity linking & 1 & 18,395 & 4,784 & 4,463 \\
        \textbf{WnWi}~\cite{wned} & Entity Linking & 1 & - & 3,396 & 3,376 \\
        \textbf{WnCw}~\cite{wned} & Entity Linking & 1 & - & 5,599 & 5,543 \\
        \textbf{T-REx}~\cite{trex} & Slot Filling & 1.26 & 2,284,168 & 5,000 & 5,000 \\
        \textbf{zsRE}~\cite{zsre} & Slot Filling & 1 & 147,909 & 3,724 & 4,966 \\
        \textbf{NQ}~\cite{nq} & Open Domain QA & 1.57 & 87,372 & 2,837 & 1,444 \\
        \textbf{HoPo}~\cite{hotpotqa} & Open Domain QA & 2 & 88,869 & 5,600 & 5,569 \\
        \textbf{TQA}~\cite{triviaqa} & Open Domain QA & 1.68 & 61,844 & 5,359 & 6,586 \\
        \textbf{ELI5}~\cite{eli5} & Open Domain QA & 1.18 & - & 1,507 & 600 \\
        \textbf{WoW}~\cite{wow} & Dialogue & 1 & 63,734 & 3,054 & 2,944 \\
        \bottomrule
    \end{tabular}
\end{table}

\subsubsection{Learning Objective}
Based on the three pre-training tasks, we can build a large number of pseudo pairs of query and document identifiers. 
All the tasks are formulated by a standard seq2seq objective, i.e., cross entropy loss, for the pre-training, 
\begin{equation*}
	\mathcal{L}=\arg \max_{\theta}\sum_{\mathcal{D}}\sum_{m} \sum_{n} \log p(w_{m,n}|w_{\leq m, <n}, q;\theta),
\end{equation*}
where $\theta$ denotes the model parameters and $\mathcal{D}$ denotes the pre-training dataset. 
All parameters are optimized by the loss $\mathcal{L}$, and the whole model is trained in an end-to-end fashion.

\section{Experimental Settings}
In this section, we introduce our experimental settings. 

\vspace*{-2mm}
\subsection{Datasets}

We first introduce the large text corpora for pre-training and eleven downstream KILT datasets.

\subsubsection{Pre-training Corpus}

We use the English Wikipedia as the pre-training corpus, since (1) Wikipedia is publicly available and easy to collect; and (2) A large collection of documents could well support our pre-training method.
English Wikipedia contains tens of millions of documents which has been widely used in many pre-training methods. 
Following~\cite{genre}, we use 2019/08/01 Wikipedia dump pre-processed by ~\citet{kilt}.

\subsubsection{Downstream Tasks}
To verify the effectiveness of our method, we conduct experiments on the KILT benchmark~\cite{kilt} with eleven datasets spanning five distinct knowledge-intensive language tasks. 
In this work, we consider the retrieval task on the KILT benchmark, in which the model should provide a set of Wikipedia pages as evidences for final prediction with respect to the input query.
Table \ref{tab:datasets} shows the overall statistics of the retrieval tasks in KILT.

\vspace*{-2mm}
\subsection{Baselines}

We adopt two types of baseline methods for comparison, including traditional IR models and model-based IR models.

\subsubsection{Traditional IR Models}
We take several representative models that are widely used for KILT tasks as the baselines, including the sparse retrieval and  dense retrieval methods. 

\begin{itemize}[leftmargin=*]
    \item \textbf{BM25}~\cite{bm25} is a highly effective retrieval model that represents the classical probabilistic retrieval model. 
    \item \textbf{TF-IDF}~\cite{drqa} is a traditional sparse vector space retrieval model that combines bigram hashing and TF-IDF matching to return relevant documents.
    \item \textbf{DPR}~\cite{dpr} is a BERT-based dual-encoder model trained with in-batch negatives and a few hard negatives selected with BM25. 
    \item \textbf{DPR+BERT}~\cite{kilt} combines a BERT-base classifier with passages returned from DPR where the query  and retrieved passages are the input. 
     \item \textbf{DPR+BART}~\cite{kilt} incorporates an explicit retrieval step in addition to the generative pre-training together with DPR and BART.
    \item \textbf{RAG}~\cite{rag} combines pre-trained parametric and non-parametric memory for generation. 
    \item \textbf{MT-DPR}~\cite{mtdpr} jointly trains a DPR model on an extensive selection of retrieval tasks. 
     \item \textbf{BLINK+flair}~\cite{kilt} combines BLINK \cite{blink} and flair \cite{flair} retrieval solution that ranks pages according to entities in the input. 
\end{itemize}

\subsubsection{Generative Retrieval Models}

Furthermore, we consider several advanced generative retrieval  baselines.  

\begin{itemize}[leftmargin=*]
    \item \textbf{BART}~\cite{bart} is a denoising autoencoder built with a Seq2Seq model that is applicable to sequence generation tasks. Following \cite{genre,gere}, we extract the query-title pairs from each downstream task and directly fine-tune the BART for generative retrieval. Specifically, we denote BART fine-tuned under three settings, i.e., zero-shot, specific-task fine-tuning and multi-task fine-tuning, as BART$_{zs}$, BART$_{ft}$ and BART$_{mt}$, respectively. Note we report the official performance of BART$_{ft}$ on KILT test data.  
    \item \textbf{T5}~\cite{t5} is a pre-trained encoder-decoder model on a multi-task mixture of unsupervised and supervised tasks. We report the official performance of fine-tuned T5 (T5$_{ft}$) on test and dev data. 
    \item \textbf{SEAL}~\cite{seal} combines an autoregressive language model with a compressed full-text substring index.  SEAL fine-tunes BART via multi-task training on all the datasets in the KILT benchmark. 
    \item \textbf{GENRE}~\cite{genre} retrieves entities by generating their unique names. GENRE takes advantage of fine-tuning BART via multi-task training on the supervised BLINK data (i.e., 9M unique triples document-mention-entity from Wikipedia) and all the data in the KILT benchmark.

\end{itemize}

\vspace*{-2mm}
\subsection{Evaluation Metrics}
To measure the retrieval performance on the downstream KILT dataset, we use R-precision (\%) as the evaluation metric, which is  suggested in the official instructions and widely used in previous works on KILT \cite{genre, kilt, mtdpr, rag, seal}. 
R-precision is calculated as $\frac{r}{R}$, where $R$ is the number of Wikipedia pages inside each provenance set and $r$ is the number of relevant pages among the top-$R$ retrieved pages. 
Besides, we also report the average performance of all the eleven datasets.   
For the zero-shot and full fine-tuning setting, we report the performance results on both the test and dev sets.
For the zero- and low-resource settings, we report the performance results on the dev sets since the KILT leaderboard\footnote{https://eval.ai/challenge/689/leaderboard} limits the frequency of the submission for test performance.

\subsection{Implementation Details}

In this section, we describe the implementation details of CorpusBrain\footnote{The code can be found at \url{https://github.com/ict-bigdatalab/CorpusBrain}}, including model architecture, pre-training process and fine-tuning process.

\subsubsection{Model Architecture}
We use the Transformer-based encoder-decoder architecture similar to BART$_{large}$ version, where the number of Transformer layers is 12, the hidden size is 1024, the feed-forward layer size is 4096 and the number of self-attention heads is 16, for both the encoder and decoder.
The total parameters is 406M.
For a fair comparison, we leverage the same architecture in the experiments for our CorpusBrain and the baseline BART.
Besides, we use sequence modeling toolkit fairseq\footnote{https://github.com/pytorch/fairseq} for the implementation of CorpusBrain.

\subsubsection{Pre-training Process}
In this work, we leverage the Wikipedia article title as the document identifier and leave other identifiers (e.g., url and hashcode) in the future work. 
Given an article, for the ISS task, we first select the lead 2 sentences, and then randomly sample up to 10 from the rest sentences set as the pseudo queries (i.e., 12 sentences in total). 
For the LPS task, we select the top 3 paragraphs (i.e., $l$=3) as the pseudo queries. 
For the output of ISS and RPS, we use the current article title joint with titles of destination Wikipedia pages linked by $o$ anchors. 
Specifically, $o$ is in [0, 1, 2, 3, 4] with the probability of [70\%, 20\%, 5\%, 3\%, 2\%], respectively.
For the HIP task, we randomly select 3 anchors from the anchor set, and then denote the anchor context as input. 
The output is the title of the destination Wikipedia page. 
In total, for each article, we construct 18 (i.e., 12+3+3) pseudo pairs.

Considering the large cost of training from scratch, we initialize the parameters of the encoder-decoder architecture from the official BART's checkpoint. 
We use a learning rate of $3e^{-5}$ and Adam optimizer with the warmup technique, where the learning rate increases over the first 10\% of batches, and then decays linearly to zero.
The label smoothing is 0.1, the weight decay is 0.01, and the gradient norm clipping is 0.1.
We train in batches of 8192 tokens on two NVIDIA Tesla V100 32GB GPUs.

\begin{table*}[t]
    \centering
     \setlength\tabcolsep{3.2pt}
    \renewcommand{\arraystretch}{0.7}
    \caption{R-precision (\%) for the page-level retrieval task on the KILT test data. \textbf{Bold} and \underline{underline} indicates the best and second model respectively. The results are reported on the KILT  leaderboard.}
    \label{tab:test}
    \begin{tabular}{llccccccccccc|c}
        \toprule
        & & \multicolumn{1}{c}{\textbf{FC}} & \multicolumn{3}{c}{\textbf{Entity Linking}} & \multicolumn{2}{c}{\textbf{Slot Filling}} & \multicolumn{4}{c}{\textbf{Open Domain QA}} & \multicolumn{1}{c}{\textbf{Dial.}} \\
        \textbf{Model Type} & \textbf{Model} & \textbf{FEV} & \textbf{AY2} & \textbf{WnWi} & \textbf{WnCw} & \textbf{T-REx} & \textbf{zsRE} & \textbf{NQ} & \textbf{HoPo} & \textbf{TQA} & \textbf{ELI5} & \textbf{WoW} & \textbf{Avg.} \\
        \midrule
        \multicolumn{14}{c}{\textit{Zero-shot}} \\
        \midrule
        \multirow{2}{*}{\begin{minipage}{0.5in}Unspervised IR Models\end{minipage}} 
        & TF-IDF~\cite{kilt} & \underline{50.9} & \underline{3.7} & \underline{0.24} & \underline{2.1} & \underline{44.7} & \underline{60.8} & \underline{28.1} & \underline{34.1} & \textbf{46.4} & \textbf{13.7} & \textbf{49.0} & \underline{30.3} \\
        & CorpusBrain$_{zs}$ & \textbf{70.38} & \textbf{5.47} & \textbf{0.56} & \textbf{7.78} & \textbf{69.48} & \textbf{94.84} & \textbf{28.25} & \textbf{44.84} & \underline{42.76} & \underline{12.17} & \underline{29.64} & \textbf{36.92} \\
        \midrule
        \multicolumn{14}{c}{\textit{Full fine-tuning}} \\
        \midrule
        \multirow{6}{*}{\begin{minipage}{0.5in}Traditional IR Models\end{minipage}} 
        & DPR~\cite{kilt} & 55.3 & 1.8 & 0.3 & 0.5 & 13.3 & 28.9 & 54.3 & 25 & 44.5 & 10.7 & 25.5 & 23.5 \\
        & DPR + BERT~\cite{kilt} & 72.9 & - & - & - & - & 40.1 & 60.7 & 25 & 43.4 & - & - & - \\
        & DPR + BART~\cite{kilt} & 55.3 & 75.5 & 45.2 & 46.9 & 13.3 & 28.9 & 54.3 & 25.0 & 44.4 & 10.7 & 25.4 & 38.6 \\
        & RAG~\cite{kilt} & 61.9 & 72.6 & 48.1 & 47.6 & 28.7 & 53.7 & 59.5 & 30.6 & 48.7 & 11.0 & 57.8 & 47.3 \\
        & BLINK + flair~\cite{kilt} & 63.7 & 81.5 & 80.2 & 68.8 & 59.6 & 78.8 & 24.5 & 46.1 & 65.6 & 9.3 & 38.2 & 56.0 \\
        & MT-DPR~\cite{mtdpr} & 74.5 & 26.5 & 4.9 & 1.9 & 69.5 & 80.9 & 59.4 & 42.9 & 61.5 & 15.5 & 41.1 & 43.5 \\
        \midrule
        \multirow{4}{*}{\begin{minipage}{0.65in}Model-based IR Models\end{minipage}}
        & T5$_{ft}$~\cite{kilt} & - & 74.0 & 47.1 & 49.3 & - & - & - & - & - & - & - & - \\
        & BART$_{ft}$~\cite{kilt} & - & 77.6 & 45.9 & 49.2 & - & - & - & - & - & - & - & - \\
        & SEAL~\cite{seal} & 81.4 & - & - & - & 62.1 & 91.6 & \textbf{63.2} & \textbf{58.8} & 68.4 & - & 57.5 & - \\
        & GENRE~\cite{genre} & \underline{83.64} & \underline{89.85} & \underline{87.44} & \textbf{71.22} & \underline{79.42} & \underline{95.81} & 60.25 & 51.27 & \underline{69.16} & \underline{15.83} & \underline{62.88} & \underline{69.71} \\
        \midrule
        Our Approach 
        & CorpusBrain$_{mt+BLINK}$ & \textbf{84.07} & \textbf{89.98} & \textbf{88.12} & \underline{70.58} & \textbf{79.98} & \textbf{98.27} & \underline{60.32} & \underline{51.80} & \textbf{70.19} & \textbf{17.50} & \textbf{64.79} & \textbf{70.51} \\
        \bottomrule
    \end{tabular}
\end{table*}

\begin{table*}[t]
    \centering
      \setlength\tabcolsep{3.2pt}
    \renewcommand{\arraystretch}{0.7}
    \caption{R-precision (\%) for the page-level retrieval task on the KILT dev data. \textbf{Bold} and \underline{underline} indicates the best and second model respectively.}
    \label{tab:dev}
    \begin{tabular}{llccccccccccc|c}
        \toprule
        & & \multicolumn{1}{c}{\textbf{FC}} & \multicolumn{3}{c}{\textbf{Entity Linking}} & \multicolumn{2}{c}{\textbf{Slot Filling}} & \multicolumn{4}{c}{\textbf{Open Domain QA}} & \multicolumn{1}{c}{\textbf{Dial.}} \\
        \textbf{Model Type} & \textbf{Model} & \textbf{FEV} & \textbf{AY2} & \textbf{WnWi} & \textbf{WnCw} & \textbf{T-REx} & \textbf{zsRE} & \textbf{NQ} & \textbf{HoPo} & \textbf{TQA} & \textbf{ELI5} & \textbf{WoW} & \textbf{Avg.} \\
        \midrule
        \multicolumn{14}{c}{\textit{Zero-shot}} \\
        \midrule
        \multirow{3}{*}{\begin{minipage}{0.5in}Unsupervised IR Models\end{minipage}} 
        & BM25~\cite{mtdpr} & \underline{50.13} & \underline{3.47} & - & - & \underline{58.6} & \underline{66.43} & \underline{25.83} & \underline{43.95} & \underline{29.44} & - & \textbf{27.5} & - \\
        & BART$_{zs}$ & 10.2 & 0.04 & \underline{0.03} & \underline{0.09} & 6.82 & 6.98 & 0.14 & 1.18 & 1.10 & \underline{0.20} & 1.38 & 2.56 \\
        & CorpusBrain$_{zs}$ & \textbf{71.69} & \textbf{5.43} & \textbf{0.53} & \textbf{6.05} & \textbf{69.44} & \textbf{90.95} & \textbf{27.88} & \textbf{45.07} & \textbf{43.40} & \textbf{9.56} & \underline{26.23} & \textbf{36.02} \\
        \midrule
        \multicolumn{14}{c}{\textit{Full fine-tuning}} \\
        \midrule
        \multirow{3}{*}{\begin{minipage}{0.5in}Traditional IR Models\end{minipage}} 
        & DPR + BART~\cite{kilt} & 55.46 & - & 44.96 & 45.70 & 13.62 & 45.60 & 54.25 & 24.62 & 45.36 & 10.32 & - & - \\
        & RAG~\cite{kilt} & 63.50 & 77.40 & 49.00 & 46.70 & 29.26 & 65.36 & 60.31 & 30.76 & 49.26 & 10.39 & 46.66 & 48.05 \\
        & MT-DPR~\cite{mtdpr} & 74.72 & 83.78 & - & - & 69.18 & 77.23 & 61.51 & 44.21 & 61.95 & - & 39.70 & - \\
        \midrule
        \multirow{4}{*}{\begin{minipage}{0.65in}Model-based IR Models\end{minipage}}
        & T5$_{ft}$~\cite{kilt} & - & 86.62 & 47.35 & 46.58 & - & - & - & - & - & - & - & -   \\
        & BART$_{ft}$ & 80.03 & 87.98 & - & - & 74.46 & 93.91 & 50.96 & 39.21 & 66.13 & - & 50.75 & - \\
        & BART$_{mt}$ & 81.92 & 89.17 & 67.58 & 62.33 & 75.18 & 91.08 & 58.62 & 48.69 & 67.64 & 12.08 & 50.98 & 64.12 \\
        & GENRE~\cite{genre} & \underline{84.68} & \underline{92.75} & \underline{87.69} & \underline{70.57} & \underline{79.68} & 94.84 & \underline{64.26} & \underline{51.82} & \underline{71.11} & \underline{13.47} & \underline{56.32} & \underline{69.74} \\
        \midrule
         \multirow{3}{*}{\begin{minipage}{0.8in}Our Approach\end{minipage}}
        & CorpusBrain$_{ft}$ & 81.77 & 90.36 & - & - & 76.90 & \textbf{98.49} & 57.67 & 50.62 & 69.25 & - & 53.60 & - \\
        & CorpusBrain$_{mt}$ & 82.06 & 90.84 & 72.26 & 66.23 & 77.62 & \underline{98.26} & 59.10 & 50.07 & 68.78 & 12.88 & 53.75 & 66.53 \\
        & CorpusBrain$_{mt+BLINK}$ & \textbf{85.03} & \textbf{92.86} & \textbf{88.64} & \textbf{71.35} & \textbf{80.22} & \textbf{98.49} & \textbf{64.61} & \textbf{52.23} & \textbf{71.71} & \textbf{14.33} & \textbf{59.72} & \textbf{70.84} \\
        \bottomrule
    \end{tabular}
\end{table*}

\subsubsection{Fine-tuning Process}

For the retrieval tasks in the downstream KILT benchmark, we first process the original data into a Seq2Seq pair format. 
Specifically, the original input (e.g., claim, text chunk and question) remains unchanged. 
For the output, we use the \texttt{[SEP]} token to concatenate the titles of several ground-truth relevant pages. 
Then, we fine-tune CorpusBrain with the processed data, with the  learning rate as $3e^{-5}$. 
For each task, We train in batches of 4096 tokens on one NVIDIA Tesla V100 32GB GPUs. 
At the inference time, we use constrained beam search with 10 beams, and maximum decoding steps of 15. 
For the entity linking sub-task, we restrict the input sequence to be at most 384 tokens cutting the left, right, or both parts of the context around a mention.
We normalize the log-probabilities by sequence length. 

In this work, we fine-tune CorpusBrain via three different strategies: (1) We fine-tune CorpusBrain on specific downstream KILT task, denoted as CorpusBrain$_{ft}$. 
(2) We fine-tune CorpusBrain via multi-task training on all KILT retrieval data following \cite{mtdpr,genre,seal}, denoted as CorpusBrain$_{mt}$. 
Note that not all dataset available in KILT have a training set as shown in Table \ref{tab:datasets}. To address all tasks, multi-task training has been a common approach for the KILT benchmark. 
(3) We follow the fine-tuning setting used in GENRE \cite{genre} to train CorpusBrain on BLINK \cite{blink} and all KILT data simultaneously, denoted as CorpusBrain$_{mt+BLINK}$.

\section{Experimental Results}
Our experiments mainly target the following research questions: 
\begin{itemize}[leftmargin=*]
\item \textbf{RQ1:} How does CorpusBrain perform compared with strong retrieval baselines across both the unsupervised and supervised evaluations? 
\item \textbf{RQ2:} What is the performance difference of CorpusBrain under specific-task fine-tuning and multi-task fine-tuning?  
\item \textbf{RQ3:} How do the three pre-training tasks of CorpusBrain  affect the retrieval performance? 
\item \textbf{RQ4:} How does CorpusBrain perform under the low-resource setting?
\item \textbf{RQ5:} How does CorpusBrain perform compared with traditional baselines in terms of memory footprint and inference time? 
\item \textbf{RQ6:} Can we better understand how different models perform via some case studies?
\end{itemize}

\vspace*{-2mm}
\subsection{Baseline Comparison}
To answer \textbf{RQ1}, we compare CorpusBrain with various strong baselines on the KILT benchmark. 
Table ~\ref{tab:test} and Table ~\ref{tab:dev} show the R-precision performance on the test and dev set, respectively. 
We can observe that: 
(1) Among the traditional IR models, the TF-IDF model performs pretty well on the test set and even outperforms the strong dense retrieval method DPR with enough supervised data. 
(2) The generative retrieval models can outperform traditional IR models significantly across all the datasets, indicating the effectiveness of integrating all the components in traditional pipelines into a single unified model. 
(3) Our CorpusBrain$_{zs}$ under zero-shot setting already achieves comparable results to most traditional baselines via full fine-tuning. 
For example, the R-precision of CorpusBrain$_{zs}$ and DPR is 94.84\% and 28.9\%, respectively, on the zsRE dataset. 


When we look at the generative retrieval baselines, we find that: (1) GENRE performs the best among these baselines in terms of most datasets, which benefits from multi-task training on supervised BLINK and entire KILT dataset at the same time using 128 GPUs.  
(2) CorpusBrain$_{ft}$ only fine-tuned on specific tasks can obtain comparable results over GENRE, which could better serve the practical use of search systems. 
CorpusBrain$_{mt+BLINK}$ outperforms GENRE on all the dev sets significantly (p-value < 0.05) and 10 out of 11 test sets. These results suggest that the pre-training tasks do help obtain a better understanding of the corpus for document retrieval. 
(3) CorpusBrain$_{mt+BLINK}$ is able to achieve the best performance among all the baselines for all 11 dev sets, while for 8 of 11 test sets and the second place for other 3 test sets. Compared with baselines which are applicable to all 11 tasks, CorpusBrain$_{mt+BLINK}$ performs the best on both the dev and test set.


Finally, the observations from the KILT leaderboard\footnote{We report the KILT leaderboard results on August 16, 2022.} with 11 KILT tasks are as follows: 
(1) As a universal retrieval model, our CorpusBrain$_{mt+BLINK}$ wins the 1st place for 3 tasks (AY2, WnWi and WoW), the 2nd place for 3 tasks (WnCw, zsRE and ELI5), and the 3rd place for 4 tasks (FEV, T-REx, HoPo and TQA). 
These results indicate the good generalization ability of our method, which is able to memorize the knowledge about the corpus. 
(2) For the leaderboard top methods, generally speaking, they are specially optimized for each task, or incorporate extra information to enhance the retrieval performance. 
For example, Re2G leverages the information in the reader component to train a separate model for each task, while the top-1 model TABi \cite{tabi} for T-REx leverages the knowledge graph, which is quite effective for the entity-related task. 
We refer readers to the leaderboard for the performance of these methods.  
(3) CorpusBrain under-performs others especially for QA datasets (e.g., NQ, HoPo, and TQA).   
The reason might be that the pseudo queries in our pre-training tasks are mainly declarative sentences, which are quite different from the questions in QA. 
For ELI5, since there is no training data, all leaderboard methods perform poorly but our CorpusBrain still achieves the 2nd. 
We will investigate to further enhance the generalization of our method.

\begin{table*}[t]
    \centering
 \renewcommand{\arraystretch}{0.78}
 \setlength\tabcolsep{6pt}
    \caption{The zero-shot R-precision (\%) of CorpusBrain with different pre-training tasks. Best results are marked bold.}
    \label{tab:tasks}
    \begin{tabular}{cccccccccccc|c}
        \toprule
        & \multicolumn{1}{c}{\textbf{FC}} & \multicolumn{3}{c}{\textbf{Entity Linking}} & \multicolumn{2}{c}{\textbf{Slot Filling}} & \multicolumn{4}{c}{\textbf{Open Domain QA}} & \multicolumn{1}{c}{\textbf{Dial.}} \\
        \textbf{Pre-training task} & \textbf{FEV} & \textbf{AY2} & \textbf{WnWi} & \textbf{WnCW} & \textbf{T-REx} & \textbf{zsRE} & \textbf{NQ} & \textbf{HoPo} & \textbf{TQA} & \textbf{ELI5} & \textbf{WoW} & \textbf{Avg.} \\
        \midrule
        ISS+LPS+HIP & \textbf{71.69} & 5.43 & 0.53 & 6.05 & \textbf{69.44} & \textbf{90.95} & \textbf{27.88} & \textbf{45.07} & \textbf{43.4} & \textbf{9.56} & \textbf{26.23} & \textbf{36.02} \\
        \midrule
        ISS & 64.89 & 4.91 & 0.65 & 5.70 & 58.8 & 79.27 & 22.91 & 44.09 & 42.10 & 7.70 & 27.14 & 32.56 \\
        LPS & 61.5 & 4.39 & 0.47 & 5.88 & 53.7 & 69.44 & 23.48 & 43.12 & 35.45 & 7.10 & 25.61 & 30.01\\
        HIP & 58.19 & \textbf{5.77} & \textbf{0.85} & \textbf{6.14} & 51.1 & 68.34 & 21.75 & 42.21 & 32.79 & 7.10 & 18.30 & 28.41 \\
        \bottomrule
    \end{tabular}
\end{table*}

\begin{figure*}[t]
 \centering
 \includegraphics[scale=0.545]{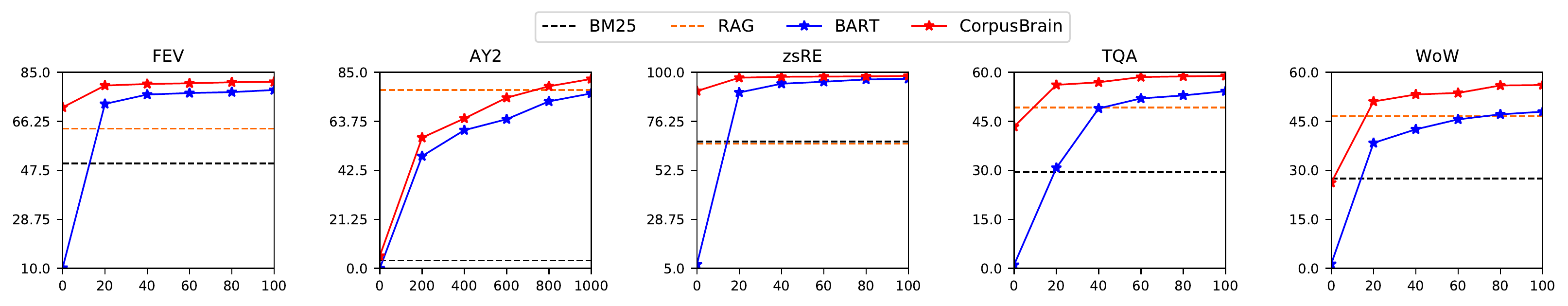}
 \caption{Fine-tuning with limited supervised data. The red and blue solid lines denote CorpusBrain and BART, respectively. The orange and gray dashed lines are RAG fine-tuned using the full supervised data and unsupervised BM25, respectively. We report the R-precision (\%) results on the dev set. }
 \label{fig:low_resource}
\end{figure*}

\vspace*{-2mm}
\subsection{Impact of Fine-tuning Strategies}
To answer \textbf{RQ2}, we further analyze the effect of different fine-tuning strategies, i.e., specific-task fine-tuning and multi-task fine-tuning, on the KILT dev set. 
We conduct a comparison of the retrieval performance between CorpusBrain and BART. 

As seen in Table ~\ref{tab:dev}, we can see that:
(1) CorpusBrain under multi-task fine-tuning achieves better results than under specific-task fine-tuning. 
This reason may be that all tasks share a common objective to generate document identifiers and boost each other in the training process.  
(2) Under both the specific-task and multi-task fine-tuning,  CorpusBrain outperforms BART over all the KILT tasks. 
This indicates that the designed three objective in CorpusBrain  which resembles the relevance relationship between the query and document identifier could contribute to the retrieval task. 
(3) CorpusBrain converges faster than BART at both fine-tuning strategies, indicating that with the encoded corpus information, CorpusBrain could easily adapt to the downstream KILT tasks via the same learning objective as in the pre-training stage.

\vspace*{-2mm}
\subsection{Impact of Pre-training Tasks}
To answer \textbf{RQ3}, we conduct a thorough ablation study on different pre-training tasks in CorpusBrain. 
Specifically, we pre-train the encoder-decoder model with ISS, LPS and HTP, respectively and evaluate the retrieval performance under the zero-shot setting on the dev set. 
For fair comparison, we set the number of pseudo (query,  document identifiers) pairs extracted from each Wikipedia article for each pre-training task as 18. 

Table~\ref{tab:tasks} shows the individual performance of three pre-training tasks. We can see that: 
(1) In general, ISS has the best performance, followed by LPS, and then HIP. 
One possible reason is that the input of 7 downstream KILT tasks is sentence-level, the (query, document identifier) pairs defined by the sampled sentences are suitable for these generative retrieval task. 
(2) HIP outperforms ISS and LPS on the entity linking tasks, i.e., AY2, WnWi and WnCw. 
This is because that the HTP task can capture the inter-page relation, which is similar to the entity linking formulation. 
Note CorpusBrain without fine-tuning performs poorly on entity linking under all the pre-training tasks. 
The reason is that we do not add any tokens (e.g., \texttt{[START\_ENT]} and \texttt{[END\_ENT]}) to denote the mention of entities at pre-training stage and thus CorpusBrain easily generates the same titles for different entites. 
(3) With all the pre-training tasks, CorpusBrain achieves the best performance.
This results demonstrate the effectiveness of all the three pre-training tasks and the importance to measure different granularities of semantics between the query and document.

\vspace*{-2mm}
\subsection{Zero- and Low-Resource Settings}

In real-world practice, it is often time-consuming and difficult to collect a large number of relevance labels to train or fine-tune a retrieval model for various KILT tasks. 
To answer \textbf{RQ4}, we simulate the low-resource retrieval setting for five datasets, i.e., FEV, AY2, zsRE, TQA and WoW, spanning five varied classes of KILT tasks. 
Following \cite{gao2021simcse}, we randomly select different   numbers of instances from the original training set for fine-tuning CorpusBrain.  
Specifically, we randomly pick 20, 40, 60, 80 and 100 instances for the FEV, zsRE, TQA and WoW dataset, and pick  200, 400, 600, 800 and 1000 instances for AY2.
We fine-tune CorpusBrain and BART on each downstream dataset,  with batches of 4096 tokens, learning rate as 3e-5, and pick the last checkpoint to evaluate the performance on the original dev set.

\begin{table*}[t]
	\caption{An example from the ELI5 dev set. The query is ``Are there any actual laws against false advertisement?'', and the titles of ground-truth relevant articles are ``False advertising'' and ``Media regulation''. We show the top-5 beams returned by GENRE, BART$_{mt}$ and CorpusBrain$_{mt+BLINK}$. Correct results are marked bold.}
	\label{tab:case}
 \renewcommand{\arraystretch}{0.9}
 \setlength\tabcolsep{8pt}
	\begin{tabular}{cccc}
		\toprule 
		\textbf{Beam} & \textbf{GENRE} & \textbf{BART$_{mt}$} & \textbf{CorpusBrain} \\ 
		\midrule
		1 & False advertising & False advertising \texttt{[SEP]} U.S. Patent No. 1 & \textbf{False advertising \texttt{[SEP]} Media regulation} \\
		2 & United States trademark law & Federal Trade Commission & False advertising \\
		3 & Anti-discrimination law & Misrepresentation \texttt{[SEP]} Ad serving & False advertising \texttt{[SEP]} Misrepresentation\\
		4 & Advertising to children & Misleading or deceptive conduct & False advertising \texttt{[SEP]} Legal liability \\
		5 & Anti-terrorism legislation & Federal Trade Commission & Advertising\\
		\bottomrule 
	\end{tabular}
\end{table*}

As shown in Figure \ref{fig:low_resource}, we can observe that:
(1) CorpusBrain outperforms BART on all the five datasets by   fine-tuning on the same limited supervised data, demonstrating that CorpusBrain is able to encode the relevance information about a given corpus through the pre-training. 
Furthermore, CorpusBrain achieves much better zero-shot performance than BART. 
(2) For the five datasets, CorpusBrain fine-tuned on limited supervised data can achieve competitive results with RAG fine-tuned on the full supervised datasets. 
For example, CorpusBrain fine-tuned with only 20 examples has outperformed RAG on TQA and WoW datasets. 
The results demonstrate that by fine-tuning with small numbers of supervised pairs, CorpusBrain is able to adapt to the target task quickly. 
(3) Under the zero resource setting, for example, CorpusBrain can outperform RAG significantly for the FEV (71.69\% vs. 63.50\%) and zsRS dataset (90.95\% vs. 65.36\%). 
(4) CorpusBrain also beats previous state-of-the-art baseline, i.e., GENRE, with limited fine-tuning examples. 
For the zsRE dataset, with just 100 examples, CorpusBrain could be fine-tuned to retrieval documents at comparable quality (i.e., R-precision = 98.28\%) to GENRE (i.e., R-precision = 94.84\%). 
It is worth noting that GENRE was trained on the full supervised 11 datasets in the KILT benchmark plus 9M BLINK data.
These results further validate that the pre-training stage encodes all the information about the corpus into the model parameters and CorpusBrain does work like an expert with a knowledgeable brain.

\vspace*{-2mm}
\subsection{Memory and Inference Efficiency}
To answer \textbf{RQ5}, we compare CorpusBrain with two representative traditional IR models, i.e., DPR and RAG, in terms of the memory footprint and inference time. 
We compare the memory footprint (disk space) required by different models. 
Here, we evaluate the end-to-end inference time of the retrieval phase in the fact checking task (i.e., FEVER dataset). 

As shown in Table~\ref{tab:memory}, we can see that: 
(1) CorpusBrain requires the least memory footprint and model parameter regardless of the size of the corpus. 
For example, CorpusBrain occupied 34 times less memory than DPR and 20 times less memory than RAG. 
It is because that CorpusBrain uses its model parameters to store all information of the corpus while traditional IR methods store dense representations for the whole corpus increased along with the increase of corpus. 
(2) CorpusBrain has a significant reduction of the inference time of document retrieval. 
As we can see, dense retrieval models like DPR need to compute relevance over all the document dense representations, while the inference process of CorpusBrain is relatively quite simple, where the inference time is directly proportional to the beam size with a limited overhead by constrained decoding. 
These results show that CorpusBrain can be well deployed in resource-limited platforms due to the memory-efficiency and time-efficiency.
 
\begin{table}[t]
	\caption{Comparisons on the memory footprint, the number of model parameters and inference time.}
	\label{tab:memory}
 \renewcommand{\arraystretch}{0.8}
 \setlength\tabcolsep{8pt}
	\begin{tabular}{lrrr}
		\toprule 
		\textbf{Model} & \textbf{Memory} & \textbf{Parameter} & \textbf{Time} \\ 
		\midrule
		DPR & 70.9GB & 220M & 14.01ms \\
		RAG & 40.4GB & 626M & 9.86ms \\
		\midrule
		\textbf{CorpusBrain} & \textbf{2.1GB} & \textbf{406M} & \textbf{5.32ms} \\ 
		\bottomrule 
	\end{tabular}
\end{table}

\vspace*{-2mm}
\subsection{Case Study}
To answer \textbf{RQ6}, we conduct some case studies to better understand how different models perform. 
We take one query from the ELI5 dev set (open-domain QA task) as an example, and show the generated Wikipedia article titles from our CorpusBrain$_{mt+BLINK}$ model as well as that from the strong baselines GENRE and BART$_{mt}$. 

As shown in Table~\ref{tab:case}, we have the following observations: 
(1) GENRE only predicts one title at each beam, but can obtain multiple titles via a post-processing way, i.e., selecting a fixed number of top-ranked beams. 
In this case, although GENRE could successfully predict one relevant document ``False advertising'', it fails to predict another relevant document in other beams. 
The reason may be that using beams to return multiple documents can not well model the dependency information between documents. 
(2) BART$_{mt}$ has the ability to model the document dependency for generating multiple titles. 
Nonetheless, without adequate pre-training tasks used for encoding the knowledge about the corpus, BART$_{mt}$ may not be able to make totally correct ground-truth documents. 
(3) CorpusBrain$_{mt+BLINK}$ can generate right document titles. 
Note in the other beams, CorpusBrain$_{mt+BLINK}$ can still generate the right title (i.e., ``false advertising'') or potential right titles (e.g., ``legal liability''). 
As we look at the ``legal liability'' article, we found that it is actually a very proper supporting article.  
These results again demonstrate the effectiveness of the proposed pre-training tasks. 

\vspace*{-2mm}
\section{Conclusion}
In this paper, we have proposed \textit{CorpusBrain}, a novel  pre-trained generative retrieval model to encode all information about the corpus into its parameters.
To train such a strong generative model, we delicately devised a set of pre-training tasks to emphasize different aspects of semantics between queries and documents. 
The key idea is to sample a context from one document as a pseudo query and generate the document identifiers of source or destination documents based on hyperlinks.  
CorpusBrain just needs to pre-train one model and could be then  adapted to improve a diversity of downstream KILT tasks without the need of constructing additional index. 
Through experiments on the KILT benchmark in terms of the retrieval task, CorpusBrain achieved significant improvements over strong baseline approaches. 
We also showed that CorpusBrain can achieve strong performance under both the zero- and low-resource settings.

In future work, we would like to explore other document identifiers, e.g., page url and HashCode, and investigate new ways to further enhance the pre-training tailored for generative retrieval. 
Besides, it is worthwhile to go beyond the search component in the KILT tasks. 
We would try to test the ability of CorpusBrain over other types of downstream IR tasks, such as ad-hoc retrieval, passage retrieval in QA or response retrieval in dialog systems. 
Furthermore, it is valuable to design an end-to-end KILT system in a fully generative way.

\begin{acks}
This work was funded by the National Natural Science Foundation of China (NSFC) under Grants No. 62006218 and 61902381, the Youth Innovation Promotion Association CAS under Grants No. 20144310, and 2021100, the Young Elite Scientist Sponsorship Program by CAST under Grants No. YESS20200121, and the Lenovo-CAS Joint Lab Youth Scientist Project.
\end{acks}

\clearpage
\bibliographystyle{ACM-Reference-Format}
\balance
\bibliography{main}


\begin{thebibliography}{52}


\ifx \showCODEN    \undefined \def \showCODEN     #1{\unskip}     \fi
\ifx \showDOI      \undefined \def \showDOI       #1{#1}\fi
\ifx \showISBNx    \undefined \def \showISBNx     #1{\unskip}     \fi
\ifx \showISBNxiii \undefined \def \showISBNxiii  #1{\unskip}     \fi
\ifx \showISSN     \undefined \def \showISSN      #1{\unskip}     \fi
\ifx \showLCCN     \undefined \def \showLCCN      #1{\unskip}     \fi
\ifx \shownote     \undefined \def \shownote      #1{#1}          \fi
\ifx \showarticletitle \undefined \def \showarticletitle #1{#1}   \fi
\ifx \showURL      \undefined \def \showURL       {\relax}        \fi
\providecommand\bibfield[2]{#2}
\providecommand\bibinfo[2]{#2}
\providecommand\natexlab[1]{#1}
\providecommand\showeprint[2][]{arXiv:#2}

\bibitem[Akbik et~al\mbox{.}(2019)]%
        {flair}
\bibfield{author}{\bibinfo{person}{Alan Akbik}, \bibinfo{person}{Tanja
  Bergmann}, \bibinfo{person}{Duncan Blythe}, \bibinfo{person}{Kashif Rasul},
  \bibinfo{person}{Stefan Schweter}, {and} \bibinfo{person}{Roland Vollgraf}.}
  \bibinfo{year}{2019}\natexlab{}.
\newblock \showarticletitle{FLAIR: An easy-to-use framework for
  state-of-the-art NLP}. In \bibinfo{booktitle}{\emph{Proceedings of the 2019
  Conference of the North American Chapter of the Association for Computational
  Linguistics (Demonstrations)}}. \bibinfo{pages}{54--59}.
\newblock


\bibitem[Bevilacqua et~al\mbox{.}(2022)]%
        {seal}
\bibfield{author}{\bibinfo{person}{Michele Bevilacqua},
  \bibinfo{person}{Giuseppe Ottaviano}, \bibinfo{person}{Patrick Lewis},
  \bibinfo{person}{Wen tau Yih}, \bibinfo{person}{Sebastian Riedel}, {and}
  \bibinfo{person}{Fabio Petroni}.} \bibinfo{year}{2022}\natexlab{}.
\newblock \showarticletitle{Autoregressive Search Engines: Generating
  Substrings as Document Identifiers}. In \bibinfo{booktitle}{\emph{arXiv
  pre-print 2204.10628}}.
\newblock
\urldef\tempurl%
\url{https://arxiv.org/abs/2204.10628}
\showURL{%
\tempurl}


\bibitem[Burges et~al\mbox{.}(2006)]%
        {burges2006learning}
\bibfield{author}{\bibinfo{person}{Christopher Burges}, \bibinfo{person}{Robert
  Ragno}, {and} \bibinfo{person}{Quoc Le}.} \bibinfo{year}{2006}\natexlab{}.
\newblock \showarticletitle{Learning to rank with nonsmooth cost functions}.
\newblock \bibinfo{journal}{\emph{NIPS}}  \bibinfo{volume}{19}
  (\bibinfo{year}{2006}).
\newblock


\bibitem[Chang et~al\mbox{.}(2019)]%
        {chang2019pre}
\bibfield{author}{\bibinfo{person}{Wei-Cheng Chang}, \bibinfo{person}{X~Yu
  Felix}, \bibinfo{person}{Yin-Wen Chang}, \bibinfo{person}{Yiming Yang}, {and}
  \bibinfo{person}{Sanjiv Kumar}.} \bibinfo{year}{2019}\natexlab{}.
\newblock \showarticletitle{Pre-training Tasks for Embedding-based Large-scale
  Retrieval}. In \bibinfo{booktitle}{\emph{ICLR}}.
\newblock


\bibitem[Chen et~al\mbox{.}(2017)]%
        {drqa}
\bibfield{author}{\bibinfo{person}{Danqi Chen}, \bibinfo{person}{Adam Fisch},
  \bibinfo{person}{Jason Weston}, {and} \bibinfo{person}{Antoine Bordes}.}
  \bibinfo{year}{2017}\natexlab{}.
\newblock \showarticletitle{Reading Wikipedia to answer open-domain questions}.
  In \bibinfo{booktitle}{\emph{55th Annual Meeting of the Association for
  Computational Linguistics, ACL 2017}}. \bibinfo{pages}{1870--1879}.
\newblock


\bibitem[Chen et~al\mbox{.}(2022)]%
        {gere}
\bibfield{author}{\bibinfo{person}{Jiangui Chen}, \bibinfo{person}{Ruqing
  Zhang}, \bibinfo{person}{Jiafeng Guo}, \bibinfo{person}{Yixing Fan}, {and}
  \bibinfo{person}{Xueqi Cheng}.} \bibinfo{year}{2022}\natexlab{}.
\newblock \showarticletitle{GERE: Generative Evidence Retrieval for Fact
  Verification}.
\newblock \bibinfo{journal}{\emph{arXiv preprint arXiv:2204.05511}}
  (\bibinfo{year}{2022}).
\newblock


\bibitem[Dai and Callan(2020)]%
        {deepct}
\bibfield{author}{\bibinfo{person}{Zhuyun Dai} {and} \bibinfo{person}{Jamie
  Callan}.} \bibinfo{year}{2020}\natexlab{}.
\newblock \showarticletitle{Context-aware term weighting for first stage
  passage retrieval}. In \bibinfo{booktitle}{\emph{SIGIR}}.
  \bibinfo{pages}{1533--1536}.
\newblock


\bibitem[Dai et~al\mbox{.}(2018)]%
        {dai2018convolutional}
\bibfield{author}{\bibinfo{person}{Zhuyun Dai}, \bibinfo{person}{Chenyan
  Xiong}, \bibinfo{person}{Jamie Callan}, {and} \bibinfo{person}{Zhiyuan Liu}.}
  \bibinfo{year}{2018}\natexlab{}.
\newblock \showarticletitle{Convolutional neural networks for soft-matching
  n-grams in ad-hoc search}. In \bibinfo{booktitle}{\emph{WSDM}}.
  \bibinfo{pages}{126--134}.
\newblock


\bibitem[De~Cao et~al\mbox{.}(2020)]%
        {genre}
\bibfield{author}{\bibinfo{person}{Nicola De~Cao}, \bibinfo{person}{Gautier
  Izacard}, \bibinfo{person}{Sebastian Riedel}, {and} \bibinfo{person}{Fabio
  Petroni}.} \bibinfo{year}{2020}\natexlab{}.
\newblock \showarticletitle{Autoregressive Entity Retrieval}. In
  \bibinfo{booktitle}{\emph{ICLR}}.
\newblock


\bibitem[Dinan et~al\mbox{.}(2018)]%
        {wow}
\bibfield{author}{\bibinfo{person}{Emily Dinan}, \bibinfo{person}{Stephen
  Roller}, \bibinfo{person}{Kurt Shuster}, \bibinfo{person}{Angela Fan},
  \bibinfo{person}{Michael Auli}, {and} \bibinfo{person}{Jason Weston}.}
  \bibinfo{year}{2018}\natexlab{}.
\newblock \showarticletitle{Wizard of Wikipedia: Knowledge-Powered
  Conversational Agents}. In \bibinfo{booktitle}{\emph{International Conference
  on Learning Representations}}.
\newblock


\bibitem[Elsahar et~al\mbox{.}(2018)]%
        {trex}
\bibfield{author}{\bibinfo{person}{Hady Elsahar}, \bibinfo{person}{Pavlos
  Vougiouklis}, \bibinfo{person}{Arslen Remaci}, \bibinfo{person}{Christophe
  Gravier}, \bibinfo{person}{Jonathon Hare}, \bibinfo{person}{Frederique
  Laforest}, {and} \bibinfo{person}{Elena Simperl}.}
  \bibinfo{year}{2018}\natexlab{}.
\newblock \showarticletitle{T-rex: A large scale alignment of natural language
  with knowledge base triples}. In \bibinfo{booktitle}{\emph{LREC}}.
\newblock


\bibitem[Fan et~al\mbox{.}(2019)]%
        {eli5}
\bibfield{author}{\bibinfo{person}{Angela Fan}, \bibinfo{person}{Yacine
  Jernite}, \bibinfo{person}{Ethan Perez}, \bibinfo{person}{David Grangier},
  \bibinfo{person}{Jason Weston}, {and} \bibinfo{person}{Michael Auli}.}
  \bibinfo{year}{2019}\natexlab{}.
\newblock \showarticletitle{{ELI}5: Long Form Question Answering}. In
  \bibinfo{booktitle}{\emph{ACL}}. \bibinfo{publisher}{Association for
  Computational Linguistics}, \bibinfo{address}{Florence, Italy},
  \bibinfo{pages}{3558--3567}.
\newblock
\urldef\tempurl%
\url{https://doi.org/10.18653/v1/P19-1346}
\showDOI{\tempurl}


\bibitem[Frej et~al\mbox{.}(2020)]%
        {frej2020learning}
\bibfield{author}{\bibinfo{person}{Jibril Frej}, \bibinfo{person}{Philippe
  Mulhem}, \bibinfo{person}{Didier Schwab}, {and} \bibinfo{person}{Jean-Pierre
  Chevallet}.} \bibinfo{year}{2020}\natexlab{}.
\newblock \showarticletitle{Learning term discrimination}. In
  \bibinfo{booktitle}{\emph{SIGIR}}. \bibinfo{pages}{1993--1996}.
\newblock


\bibitem[Gao et~al\mbox{.}(2021)]%
        {gao2021simcse}
\bibfield{author}{\bibinfo{person}{Tianyu Gao}, \bibinfo{person}{Xingcheng
  Yao}, {and} \bibinfo{person}{Danqi Chen}.} \bibinfo{year}{2021}\natexlab{}.
\newblock \showarticletitle{SimCSE: Simple Contrastive Learning of Sentence
  Embeddings}. In \bibinfo{booktitle}{\emph{EMNLP}}.
  \bibinfo{pages}{6894--6910}.
\newblock


\bibitem[Guo et~al\mbox{.}(2022)]%
        {guo2022semantic}
\bibfield{author}{\bibinfo{person}{Jiafeng Guo}, \bibinfo{person}{Yinqiong
  Cai}, \bibinfo{person}{Yixing Fan}, \bibinfo{person}{Fei Sun},
  \bibinfo{person}{Ruqing Zhang}, {and} \bibinfo{person}{Xueqi Cheng}.}
  \bibinfo{year}{2022}\natexlab{}.
\newblock \showarticletitle{Semantic models for the first-stage retrieval: A
  comprehensive review}.
\newblock \bibinfo{journal}{\emph{TOIS}} \bibinfo{volume}{40},
  \bibinfo{number}{4} (\bibinfo{year}{2022}), \bibinfo{pages}{1--42}.
\newblock


\bibitem[Guo et~al\mbox{.}(2016)]%
        {guo2016deep}
\bibfield{author}{\bibinfo{person}{Jiafeng Guo}, \bibinfo{person}{Yixing Fan},
  \bibinfo{person}{Qingyao Ai}, {and} \bibinfo{person}{W~Bruce Croft}.}
  \bibinfo{year}{2016}\natexlab{}.
\newblock \showarticletitle{A deep relevance matching model for ad-hoc
  retrieval}. In \bibinfo{booktitle}{\emph{CIKM}}. \bibinfo{pages}{55--64}.
\newblock


\bibitem[Guo and Barbosa(2018)]%
        {wned}
\bibfield{author}{\bibinfo{person}{Zhaochen Guo} {and}
  \bibinfo{person}{Denilson Barbosa}.} \bibinfo{year}{2018}\natexlab{}.
\newblock \showarticletitle{Robust named entity disambiguation with random
  walks}.
\newblock \bibinfo{journal}{\emph{Semantic Web}} \bibinfo{volume}{9},
  \bibinfo{number}{4} (\bibinfo{year}{2018}), \bibinfo{pages}{459--479}.
\newblock


\bibitem[Hoffart et~al\mbox{.}(2011)]%
        {aida}
\bibfield{author}{\bibinfo{person}{Johannes Hoffart},
  \bibinfo{person}{Mohamed~Amir Yosef}, \bibinfo{person}{Ilaria Bordino},
  \bibinfo{person}{Hagen F{\"u}rstenau}, \bibinfo{person}{Manfred Pinkal},
  \bibinfo{person}{Marc Spaniol}, \bibinfo{person}{Bilyana Taneva},
  \bibinfo{person}{Stefan Thater}, {and} \bibinfo{person}{Gerhard Weikum}.}
  \bibinfo{year}{2011}\natexlab{}.
\newblock \showarticletitle{Robust disambiguation of named entities in text}.
  In \bibinfo{booktitle}{\emph{EMNLP}}. \bibinfo{pages}{782--792}.
\newblock


\bibitem[Hofst{\"a}tter et~al\mbox{.}(2021)]%
        {hofstatter2021efficiently}
\bibfield{author}{\bibinfo{person}{Sebastian Hofst{\"a}tter},
  \bibinfo{person}{Sheng-Chieh Lin}, \bibinfo{person}{Jheng-Hong Yang},
  \bibinfo{person}{Jimmy Lin}, {and} \bibinfo{person}{Allan Hanbury}.}
  \bibinfo{year}{2021}\natexlab{}.
\newblock \showarticletitle{Efficiently teaching an effective dense retriever
  with balanced topic aware sampling}. In \bibinfo{booktitle}{\emph{SIGIR}}.
  \bibinfo{pages}{113--122}.
\newblock


\bibitem[Joshi et~al\mbox{.}(2017)]%
        {triviaqa}
\bibfield{author}{\bibinfo{person}{Mandar Joshi}, \bibinfo{person}{Eunsol
  Choi}, \bibinfo{person}{Daniel Weld}, {and} \bibinfo{person}{Luke
  Zettlemoyer}.} \bibinfo{year}{2017}\natexlab{}.
\newblock \showarticletitle{{T}rivia{QA}: A Large Scale Distantly Supervised
  Challenge Dataset for Reading Comprehension}. In
  \bibinfo{booktitle}{\emph{ACL}}. \bibinfo{publisher}{Association for
  Computational Linguistics}, \bibinfo{address}{Vancouver, Canada},
  \bibinfo{pages}{1601--1611}.
\newblock


\bibitem[Karpukhin et~al\mbox{.}(2020)]%
        {dpr}
\bibfield{author}{\bibinfo{person}{Vladimir Karpukhin}, \bibinfo{person}{Barlas
  Oguz}, \bibinfo{person}{Sewon Min}, \bibinfo{person}{Patrick Lewis},
  \bibinfo{person}{Ledell Wu}, \bibinfo{person}{Sergey Edunov},
  \bibinfo{person}{Danqi Chen}, {and} \bibinfo{person}{Wen-tau Yih}.}
  \bibinfo{year}{2020}\natexlab{}.
\newblock \showarticletitle{Dense Passage Retrieval for Open-Domain Question
  Answering}. In \bibinfo{booktitle}{\emph{EMNLP}}.
  \bibinfo{pages}{6769--6781}.
\newblock


\bibitem[Kenton and Toutanova(2019)]%
        {bert}
\bibfield{author}{\bibinfo{person}{Jacob Devlin Ming-Wei~Chang Kenton} {and}
  \bibinfo{person}{Lee~Kristina Toutanova}.} \bibinfo{year}{2019}\natexlab{}.
\newblock \showarticletitle{BERT: Pre-training of Deep Bidirectional
  Transformers for Language Understanding}. In
  \bibinfo{booktitle}{\emph{NAACL-HLT}}. \bibinfo{pages}{4171--4186}.
\newblock


\bibitem[Khattab and Zaharia(2020)]%
        {khattab2020colbert}
\bibfield{author}{\bibinfo{person}{Omar Khattab} {and} \bibinfo{person}{Matei
  Zaharia}.} \bibinfo{year}{2020}\natexlab{}.
\newblock \showarticletitle{Colbert: Efficient and effective passage search via
  contextualized late interaction over bert}. In
  \bibinfo{booktitle}{\emph{SIGIR}}. \bibinfo{pages}{39--48}.
\newblock


\bibitem[Kwiatkowski et~al\mbox{.}(2019)]%
        {nq}
\bibfield{author}{\bibinfo{person}{Tom Kwiatkowski},
  \bibinfo{person}{Jennimaria Palomaki}, \bibinfo{person}{Olivia Redfield},
  \bibinfo{person}{Michael Collins}, \bibinfo{person}{Ankur Parikh},
  \bibinfo{person}{Chris Alberti}, \bibinfo{person}{Danielle Epstein},
  \bibinfo{person}{Illia Polosukhin}, \bibinfo{person}{Jacob Devlin},
  \bibinfo{person}{Kenton Lee}, {et~al\mbox{.}}}
  \bibinfo{year}{2019}\natexlab{}.
\newblock \showarticletitle{Natural questions: a benchmark for question
  answering research}.
\newblock \bibinfo{journal}{\emph{Transactions of the Association for
  Computational Linguistics}}  \bibinfo{volume}{7} (\bibinfo{year}{2019}),
  \bibinfo{pages}{453--466}.
\newblock


\bibitem[Lee et~al\mbox{.}(2019)]%
        {lee2019latent}
\bibfield{author}{\bibinfo{person}{Kenton Lee}, \bibinfo{person}{Ming-Wei
  Chang}, {and} \bibinfo{person}{Kristina Toutanova}.}
  \bibinfo{year}{2019}\natexlab{}.
\newblock \showarticletitle{Latent Retrieval for Weakly Supervised Open Domain
  Question Answering}. In \bibinfo{booktitle}{\emph{ACL}}.
  \bibinfo{pages}{6086--6096}.
\newblock


\bibitem[Leszczynski et~al\mbox{.}(2022)]%
        {tabi}
\bibfield{author}{\bibinfo{person}{Megan Leszczynski},
  \bibinfo{person}{Daniel~Y Fu}, \bibinfo{person}{Mayee~F Chen}, {and}
  \bibinfo{person}{Christopher R{\'e}}.} \bibinfo{year}{2022}\natexlab{}.
\newblock \showarticletitle{TABi: Type-Aware Bi-Encoders for Open-Domain Entity
  Retrieval}.
\newblock \bibinfo{journal}{\emph{arXiv preprint arXiv:2204.08173}}
  (\bibinfo{year}{2022}).
\newblock


\bibitem[Levy et~al\mbox{.}(2017)]%
        {zsre}
\bibfield{author}{\bibinfo{person}{Omer Levy}, \bibinfo{person}{Minjoon Seo},
  \bibinfo{person}{Eunsol Choi}, {and} \bibinfo{person}{Luke Zettlemoyer}.}
  \bibinfo{year}{2017}\natexlab{}.
\newblock \showarticletitle{Zero-Shot Relation Extraction via Reading
  Comprehension}. In \bibinfo{booktitle}{\emph{CoNLL}}.
  \bibinfo{pages}{333--342}.
\newblock


\bibitem[Lewis et~al\mbox{.}(2020a)]%
        {bart}
\bibfield{author}{\bibinfo{person}{Mike Lewis}, \bibinfo{person}{Yinhan Liu},
  \bibinfo{person}{Naman Goyal}, \bibinfo{person}{Marjan Ghazvininejad},
  \bibinfo{person}{Abdelrahman Mohamed}, \bibinfo{person}{Omer Levy},
  \bibinfo{person}{Veselin Stoyanov}, {and} \bibinfo{person}{Luke
  Zettlemoyer}.} \bibinfo{year}{2020}\natexlab{a}.
\newblock \showarticletitle{BART: Denoising Sequence-to-Sequence Pre-training
  for Natural Language Generation, Translation, and Comprehension}. In
  \bibinfo{booktitle}{\emph{ACL}}. \bibinfo{pages}{7871--7880}.
\newblock


\bibitem[Lewis et~al\mbox{.}(2020b)]%
        {rag}
\bibfield{author}{\bibinfo{person}{Patrick Lewis}, \bibinfo{person}{Ethan
  Perez}, \bibinfo{person}{Aleksandra Piktus}, \bibinfo{person}{Fabio Petroni},
  \bibinfo{person}{Vladimir Karpukhin}, \bibinfo{person}{Naman Goyal},
  \bibinfo{person}{Heinrich K{\"u}ttler}, \bibinfo{person}{Mike Lewis},
  \bibinfo{person}{Wen-tau Yih}, \bibinfo{person}{Tim Rockt{\"a}schel},
  {et~al\mbox{.}}} \bibinfo{year}{2020}\natexlab{b}.
\newblock \showarticletitle{Retrieval-augmented generation for
  knowledge-intensive nlp tasks}.
\newblock \bibinfo{journal}{\emph{NIPS}}  \bibinfo{volume}{33}
  (\bibinfo{year}{2020}), \bibinfo{pages}{9459--9474}.
\newblock


\bibitem[Li(2014)]%
        {li2014learning}
\bibfield{author}{\bibinfo{person}{Hang Li}.} \bibinfo{year}{2014}\natexlab{}.
\newblock \showarticletitle{Learning to rank for information retrieval and
  natural language processing}.
\newblock \bibinfo{journal}{\emph{Synthesis lectures on human language
  technologies}} \bibinfo{volume}{7}, \bibinfo{number}{3}
  (\bibinfo{year}{2014}), \bibinfo{pages}{1--121}.
\newblock


\bibitem[Liu et~al\mbox{.}(2009)]%
        {liu2009learning}
\bibfield{author}{\bibinfo{person}{Tie-Yan Liu} {et~al\mbox{.}}}
  \bibinfo{year}{2009}\natexlab{}.
\newblock \showarticletitle{Learning to rank for information retrieval}.
\newblock \bibinfo{journal}{\emph{Foundations and Trends{\textregistered} in
  Information Retrieval}} \bibinfo{volume}{3}, \bibinfo{number}{3}
  (\bibinfo{year}{2009}), \bibinfo{pages}{225--331}.
\newblock


\bibitem[Ma et~al\mbox{.}(2022)]%
        {ma2022pre}
\bibfield{author}{\bibinfo{person}{Xinyu Ma}, \bibinfo{person}{Jiafeng Guo},
  \bibinfo{person}{Ruqing Zhang}, \bibinfo{person}{Yixing Fan}, {and}
  \bibinfo{person}{Xueqi Cheng}.} \bibinfo{year}{2022}\natexlab{}.
\newblock \showarticletitle{Pre-train a Discriminative Text Encoder for Dense
  Retrieval via Contrastive Span Prediction}.
\newblock \bibinfo{journal}{\emph{arXiv preprint arXiv:2204.10641}}
  (\bibinfo{year}{2022}).
\newblock


\bibitem[Ma et~al\mbox{.}(2021b)]%
        {prop}
\bibfield{author}{\bibinfo{person}{Xinyu Ma}, \bibinfo{person}{Jiafeng Guo},
  \bibinfo{person}{Ruqing Zhang}, \bibinfo{person}{Yixing Fan},
  \bibinfo{person}{Xiang Ji}, {and} \bibinfo{person}{Xueqi Cheng}.}
  \bibinfo{year}{2021}\natexlab{b}.
\newblock \showarticletitle{Prop: Pre-training with representative words
  prediction for ad-hoc retrieval}. In \bibinfo{booktitle}{\emph{WSDM}}.
  \bibinfo{pages}{283--291}.
\newblock


\bibitem[Ma et~al\mbox{.}(2021c)]%
        {bprop}
\bibfield{author}{\bibinfo{person}{Xinyu Ma}, \bibinfo{person}{Jiafeng Guo},
  \bibinfo{person}{Ruqing Zhang}, \bibinfo{person}{Yixing Fan},
  \bibinfo{person}{Yingyan Li}, {and} \bibinfo{person}{Xueqi Cheng}.}
  \bibinfo{year}{2021}\natexlab{c}.
\newblock \showarticletitle{B-PROP: bootstrapped pre-training with
  representative words prediction for ad-hoc retrieval}. In
  \bibinfo{booktitle}{\emph{SIGIR}}. \bibinfo{pages}{1513--1522}.
\newblock


\bibitem[Ma et~al\mbox{.}(2021a)]%
        {ma2021pre}
\bibfield{author}{\bibinfo{person}{Zhengyi Ma}, \bibinfo{person}{Zhicheng Dou},
  \bibinfo{person}{Wei Xu}, \bibinfo{person}{Xinyu Zhang}, \bibinfo{person}{Hao
  Jiang}, \bibinfo{person}{Zhao Cao}, {and} \bibinfo{person}{Ji-Rong Wen}.}
  \bibinfo{year}{2021}\natexlab{a}.
\newblock \showarticletitle{Pre-training for Ad-hoc Retrieval: Hyperlink is
  Also You Need}. In \bibinfo{booktitle}{\emph{CIKM}}.
  \bibinfo{pages}{1212--1221}.
\newblock


\bibitem[Maillard et~al\mbox{.}(2021)]%
        {mtdpr}
\bibfield{author}{\bibinfo{person}{Jean Maillard}, \bibinfo{person}{Vladimir
  Karpukhin}, \bibinfo{person}{Fabio Petroni}, \bibinfo{person}{Wen-tau Yih},
  \bibinfo{person}{Barlas Oguz}, \bibinfo{person}{Veselin Stoyanov}, {and}
  \bibinfo{person}{Gargi Ghosh}.} \bibinfo{year}{2021}\natexlab{}.
\newblock \showarticletitle{Multi-Task Retrieval for Knowledge-Intensive
  Tasks}. In \bibinfo{booktitle}{\emph{ACL}}. \bibinfo{pages}{1098--1111}.
\newblock


\bibitem[Metzler et~al\mbox{.}(2021)]%
        {metzler2021rethinking}
\bibfield{author}{\bibinfo{person}{Donald Metzler}, \bibinfo{person}{Yi Tay},
  \bibinfo{person}{Dara Bahri}, {and} \bibinfo{person}{Marc Najork}.}
  \bibinfo{year}{2021}\natexlab{}.
\newblock \showarticletitle{Rethinking search: making domain experts out of
  dilettantes}. In \bibinfo{booktitle}{\emph{ACM SIGIR Forum}},
  Vol.~\bibinfo{volume}{55}. ACM New York, NY, USA, \bibinfo{pages}{1--27}.
\newblock


\bibitem[Petroni et~al\mbox{.}(2021)]%
        {kilt}
\bibfield{author}{\bibinfo{person}{Fabio Petroni}, \bibinfo{person}{Aleksandra
  Piktus}, \bibinfo{person}{Angela Fan}, \bibinfo{person}{Patrick Lewis},
  \bibinfo{person}{Majid Yazdani}, \bibinfo{person}{Nicola De~Cao},
  \bibinfo{person}{James Thorne}, \bibinfo{person}{Yacine Jernite},
  \bibinfo{person}{Vladimir Karpukhin}, \bibinfo{person}{Jean Maillard},
  \bibinfo{person}{Vassilis Plachouras}, \bibinfo{person}{Tim Rockt{\"a}schel},
  {and} \bibinfo{person}{Sebastian Riedel}.} \bibinfo{year}{2021}\natexlab{}.
\newblock \showarticletitle{{KILT}: a Benchmark for Knowledge Intensive
  Language Tasks}. In \bibinfo{booktitle}{\emph{Proceedings of the 2021
  Conference of the North American Chapter of the Association for Computational
  Linguistics: Human Language Technologies}}. \bibinfo{publisher}{Association
  for Computational Linguistics}, \bibinfo{address}{Online},
  \bibinfo{pages}{2523--2544}.
\newblock
\urldef\tempurl%
\url{https://doi.org/10.18653/v1/2021.naacl-main.200}
\showDOI{\tempurl}


\bibitem[Raffel et~al\mbox{.}(2020)]%
        {t5}
\bibfield{author}{\bibinfo{person}{Colin Raffel}, \bibinfo{person}{Noam
  Shazeer}, \bibinfo{person}{Adam Roberts}, \bibinfo{person}{Katherine Lee},
  \bibinfo{person}{Sharan Narang}, \bibinfo{person}{Michael Matena},
  \bibinfo{person}{Yanqi Zhou}, \bibinfo{person}{Wei Li}, {and}
  \bibinfo{person}{Peter~J Liu}.} \bibinfo{year}{2020}\natexlab{}.
\newblock \showarticletitle{Exploring the Limits of Transfer Learning with a
  Unified Text-to-Text Transformer}.
\newblock \bibinfo{journal}{\emph{Journal of Machine Learning Research}}
  \bibinfo{volume}{21} (\bibinfo{year}{2020}), \bibinfo{pages}{1--67}.
\newblock


\bibitem[Robertson and Zaragoza(2009)]%
        {bm25}
\bibfield{author}{\bibinfo{person}{Stephen Robertson} {and}
  \bibinfo{person}{Hugo Zaragoza}.} \bibinfo{year}{2009}\natexlab{}.
\newblock \bibinfo{booktitle}{\emph{The probabilistic relevance framework: BM25
  and beyond}}.
\newblock \bibinfo{publisher}{Now Publishers Inc}.
\newblock


\bibitem[Robertson and Jones(1976)]%
        {robertson1976relevance}
\bibfield{author}{\bibinfo{person}{Stephen~E Robertson} {and}
  \bibinfo{person}{K~Sparck Jones}.} \bibinfo{year}{1976}\natexlab{}.
\newblock \showarticletitle{Relevance weighting of search terms}.
\newblock \bibinfo{journal}{\emph{Journal of the American Society for
  Information science}} \bibinfo{volume}{27}, \bibinfo{number}{3}
  (\bibinfo{year}{1976}), \bibinfo{pages}{129--146}.
\newblock


\bibitem[Salton et~al\mbox{.}(1975)]%
        {salton1975vector}
\bibfield{author}{\bibinfo{person}{Gerard Salton}, \bibinfo{person}{Anita
  Wong}, {and} \bibinfo{person}{Chung-Shu Yang}.}
  \bibinfo{year}{1975}\natexlab{}.
\newblock \showarticletitle{A vector space model for automatic indexing}.
\newblock \bibinfo{journal}{\emph{Commun. ACM}} \bibinfo{volume}{18},
  \bibinfo{number}{11} (\bibinfo{year}{1975}), \bibinfo{pages}{613--620}.
\newblock


\bibitem[Sutskever et~al\mbox{.}(2011)]%
        {sutskever2011generating}
\bibfield{author}{\bibinfo{person}{Ilya Sutskever}, \bibinfo{person}{James
  Martens}, {and} \bibinfo{person}{Geoffrey~E Hinton}.}
  \bibinfo{year}{2011}\natexlab{}.
\newblock \showarticletitle{Generating text with recurrent neural networks}. In
  \bibinfo{booktitle}{\emph{ICML}}.
\newblock


\bibitem[Tay et~al\mbox{.}(2022)]%
        {tay2022transformer}
\bibfield{author}{\bibinfo{person}{Yi Tay}, \bibinfo{person}{Vinh~Q Tran},
  \bibinfo{person}{Mostafa Dehghani}, \bibinfo{person}{Jianmo Ni},
  \bibinfo{person}{Dara Bahri}, \bibinfo{person}{Harsh Mehta},
  \bibinfo{person}{Zhen Qin}, \bibinfo{person}{Kai Hui}, \bibinfo{person}{Zhe
  Zhao}, \bibinfo{person}{Jai Gupta}, {et~al\mbox{.}}}
  \bibinfo{year}{2022}\natexlab{}.
\newblock \showarticletitle{Transformer memory as a differentiable search
  index}.
\newblock \bibinfo{journal}{\emph{arXiv preprint arXiv:2202.06991}}
  (\bibinfo{year}{2022}).
\newblock


\bibitem[Thorne et~al\mbox{.}(2018)]%
        {fever}
\bibfield{author}{\bibinfo{person}{James Thorne}, \bibinfo{person}{Andreas
  Vlachos}, \bibinfo{person}{Christos Christodoulopoulos}, {and}
  \bibinfo{person}{Arpit Mittal}.} \bibinfo{year}{2018}\natexlab{}.
\newblock \showarticletitle{FEVER: a Large-scale Dataset for Fact Extraction
  and VERification}. In \bibinfo{booktitle}{\emph{Proceedings of the 2018
  Conference of the North American Chapter of the Association for Computational
  Linguistics: Human Language Technologies, Volume 1 (Long Papers)}}.
  \bibinfo{pages}{809--819}.
\newblock


\bibitem[Wu et~al\mbox{.}(2020)]%
        {blink}
\bibfield{author}{\bibinfo{person}{Ledell Wu}, \bibinfo{person}{Fabio Petroni},
  \bibinfo{person}{Martin Josifoski}, \bibinfo{person}{Sebastian Riedel}, {and}
  \bibinfo{person}{Luke Zettlemoyer}.} \bibinfo{year}{2020}\natexlab{}.
\newblock \showarticletitle{Scalable Zero-shot Entity Linking with Dense Entity
  Retrieval}. In \bibinfo{booktitle}{\emph{EMNLP}}.
  \bibinfo{pages}{6397--6407}.
\newblock


\bibitem[Xiong et~al\mbox{.}(2020)]%
        {xiong2020approximate}
\bibfield{author}{\bibinfo{person}{Lee Xiong}, \bibinfo{person}{Chenyan Xiong},
  \bibinfo{person}{Ye Li}, \bibinfo{person}{Kwok-Fung Tang},
  \bibinfo{person}{Jialin Liu}, \bibinfo{person}{Paul~N Bennett},
  \bibinfo{person}{Junaid Ahmed}, {and} \bibinfo{person}{Arnold Overwijk}.}
  \bibinfo{year}{2020}\natexlab{}.
\newblock \showarticletitle{Approximate Nearest Neighbor Negative Contrastive
  Learning for Dense Text Retrieval}. In \bibinfo{booktitle}{\emph{ICLR}}.
\newblock


\bibitem[Yang et~al\mbox{.}(2018)]%
        {hotpotqa}
\bibfield{author}{\bibinfo{person}{Zhilin Yang}, \bibinfo{person}{Peng Qi},
  \bibinfo{person}{Saizheng Zhang}, \bibinfo{person}{Yoshua Bengio},
  \bibinfo{person}{William Cohen}, \bibinfo{person}{Ruslan Salakhutdinov},
  {and} \bibinfo{person}{Christopher~D. Manning}.}
  \bibinfo{year}{2018}\natexlab{}.
\newblock \showarticletitle{{H}otpot{QA}: A Dataset for Diverse, Explainable
  Multi-hop Question Answering}. In \bibinfo{booktitle}{\emph{EMNLP}}.
  \bibinfo{publisher}{Association for Computational Linguistics},
  \bibinfo{address}{Brussels, Belgium}, \bibinfo{pages}{2369--2380}.
\newblock
\urldef\tempurl%
\url{https://doi.org/10.18653/v1/D18-1259}
\showDOI{\tempurl}


\bibitem[Zhan et~al\mbox{.}(2021)]%
        {zhan2021optimizing}
\bibfield{author}{\bibinfo{person}{Jingtao Zhan}, \bibinfo{person}{Jiaxin Mao},
  \bibinfo{person}{Yiqun Liu}, \bibinfo{person}{Jiafeng Guo},
  \bibinfo{person}{Min Zhang}, {and} \bibinfo{person}{Shaoping Ma}.}
  \bibinfo{year}{2021}\natexlab{}.
\newblock \showarticletitle{Optimizing dense retrieval model training with hard
  negatives}. In \bibinfo{booktitle}{\emph{SIGIR}}.
  \bibinfo{pages}{1503--1512}.
\newblock


\bibitem[Zhang et~al\mbox{.}(2020)]%
        {zhang2020pegasus}
\bibfield{author}{\bibinfo{person}{Jingqing Zhang}, \bibinfo{person}{Yao Zhao},
  \bibinfo{person}{Mohammad Saleh}, {and} \bibinfo{person}{Peter Liu}.}
  \bibinfo{year}{2020}\natexlab{}.
\newblock \showarticletitle{Pegasus: Pre-training with extracted gap-sentences
  for abstractive summarization}. In \bibinfo{booktitle}{\emph{International
  Conference on Machine Learning}}. PMLR, \bibinfo{pages}{11328--11339}.
\newblock


\bibitem[Zheng and Callan(2015)]%
        {zheng2015learning}
\bibfield{author}{\bibinfo{person}{Guoqing Zheng} {and} \bibinfo{person}{Jamie
  Callan}.} \bibinfo{year}{2015}\natexlab{}.
\newblock \showarticletitle{Learning to reweight terms with distributed
  representations}. In \bibinfo{booktitle}{\emph{SIGIR}}.
  \bibinfo{pages}{575--584}.
\newblock


\bibitem[Zhou et~al\mbox{.}(2022)]%
        {zhou2022dynamicretriever}
\bibfield{author}{\bibinfo{person}{Yujia Zhou}, \bibinfo{person}{Jing Yao},
  \bibinfo{person}{Zhicheng Dou}, \bibinfo{person}{Ledell Wu}, {and}
  \bibinfo{person}{Ji-Rong Wen}.} \bibinfo{year}{2022}\natexlab{}.
\newblock \showarticletitle{DynamicRetriever: A Pre-training Model-based IR
  System with Neither Sparse nor Dense Index}.
\newblock \bibinfo{journal}{\emph{arXiv preprint arXiv:2203.00537}}
  (\bibinfo{year}{2022}).
\newblock


\end{thebibliography}


\end{document}